\documentclass[preprint,12pt]{elsarticle}

\usepackage{algorithm}
\usepackage{algpseudocode}

\usepackage{amsmath, amssymb, amsfonts}

\usepackage{algorithm}

\usepackage{booktabs}      
\usepackage{multirow}      
\usepackage{multicol}      
\usepackage{array}         
\usepackage{tabularx}      
\usepackage{longtable}     
\usepackage{threeparttable}
\usepackage{colortbl}      
\usepackage{xcolor}        
\usepackage{booktabs}     
\usepackage{threeparttable} 
\usepackage{graphicx}
\usepackage{subcaption}
\usepackage{tikz}
\usepackage{adjustbox}

\usepackage[utf8]{inputenc}
\usepackage[T1]{fontenc}

\usepackage{hyperref}
\usepackage{url}
\usepackage{etoolbox}
\usepackage{siunitx}
\usepackage{epstopdf}
\usepackage{rotating}
\usepackage{pdfpages}
\usepackage{geometry}

\journal{Nuclear Physics B}

\begin{document}

\begin{frontmatter}



\title {ATMS-KD: Adaptive Temperature and Mixed Sample Knowledge Distillation for a Lightweight Residual CNN in Agricultural Embedded Systems}


\author[1]{Mohamed Ohamouddou}
\ead{mohamed.ohamouddou@etu.uae.ac.ma} 

\author[2]{Said Ohamouddou}
\author[2]{Abdellatif El Afia} 
\author[1]{Rafik Lasri} 
\affiliation[1]{organization={TEDAEEP Research Group, FPL, Abdelmalek saadi University},
         postcode={Quartier Mhneche II, Avenue 9 Avril B.P.2117}, 
            city={Tetouan},
            country={Morocco}}
\affiliation[2]{organization={Smart System Laboratory (SSL), ENSIAS, Mohammed V University},
           city={Rabat},
         country={Morocco}}

 

\begin{abstract}

Agricultural embedded systems require efficient deep learning models for real-time crop monitoring and maturity assessment. This study proposes ATMS-KD (Adaptive Temperature and Mixed-Sample Knowledge Distillation), a novel framework for developing lightweight CNN models suitable for resource-constrained agricultural environments. The framework combines adaptive temperature scheduling with mixed-sample augmentation to transfer knowledge from a MobileNetV3 Large teacher model (5.7M parameters) to lightweight residual CNN students. Three student configurations were evaluated: Compact (0.75× width, 1.3M parameters), Standard (1.0× width, 2.4M parameters), and Enhanced (1.25× width, 3.8M parameters). The dataset used in this study consists of images of Rosa damascena (Damask rose) collected from agricultural fields in the Dades Oasis, southeastern Morocco, providing a realistic benchmark for agricultural computer vision applications under diverse environmental conditions. Experimental evaluation on the Damascena rose maturity classification dataset demonstrated significant improvements over direct training methods. All student models achieved validation accuracies exceeding 96.7\% with ATMS-KD compared to 95-96\% with direct training. The framework outperformed eleven established knowledge distillation methods, achieving 97.11\% accuracy with the compact model—a 1.60 percentage point improvement over the second-best approach while maintaining the lowest inference latency of 72.19 ms. Knowledge retention rates exceeded 99\% for all configurations, demonstrating effective knowledge transfer regardless of student model capacity. The compact model achieved optimal computational efficiency with 13.9 samples per second throughput, meeting real-time requirements for embedded agricultural systems. Qualitative analysis confirmed robust performance across diverse field conditions including varying lighting, background complexity, and environmental factors. The proposed framework provides an effective solution for deploying accurate and efficient deep learning models in precision agriculture applications while addressing the computational constraints of agricultural embedded systems.

\end{abstract}






\end{frontmatter}


\section{Introduction}
\label{sec:introduction}

Agricultural systems worldwide face increasing pressure to enhance productivity while maintaining sustainability in the context of global population growth and climate change. Precision agriculture technologies, particularly computer vision-based crop monitoring systems, offer promising solutions for optimizing agricultural practices through real-time assessment of crop conditions \cite{kamilaris_deep_2018}. However, the deployment of sophisticated deep learning models in agricultural environments presents significant challenges due to the computational constraints of field-deployed embedded systems and the need for accurate, real-time decision-making capabilities \cite{patricio_computer_2018}.\\
Deep convolutional neural networks have demonstrated remarkable success in agricultural applications, including crop classification, disease detection, and maturity assessment \cite{saleem_plant_2019}. Recent advances in mobile-optimized architectures such as MobileNets \cite{howard_mobilenets_2017} and EfficientNets \cite{tan_efficientnet_2020} and YOLO \cite{fatehi_enhancing_2025} have reduced computational requirements while maintaining competitive accuracy levels. Nevertheless, these models often remain too resource-intensive for deployment on low-power agricultural embedded systems typically used in field monitoring applications, where factors such as battery life, processing capabilities, and environmental durability are critical considerations \cite{liakos_machine_2018}.\\
Knowledge distillation has emerged as a prominent technique for model compression, enabling the transfer of knowledge from large, complex teacher models to smaller, more efficient student networks \cite{hinton_distilling_2015}. In \cite{hussain_gradual_2024} authors proposed a progressive knowledge transfer approach that systematically distilled hash code knowledge through incremental learning stages for improved large-scale image retrieval performance. The fundamental principle involves training a compact student model to mimic the behavior of a pre-trained teacher model, typically through the use of soft targets that provide richer information than traditional hard labels \cite{gou_knowledge_2021}. This approach has shown considerable promise in reducing model size and computational requirements while preserving classification performance across various domains \cite{wang_knowledge_2022,xu_knowledge_2020}.\\
Computer vision applications in agriculture have experienced rapid development in recent years. Kamilaris and Prenafeta-Boldú \cite{kamilaris_deep_2018} provided a comprehensive review of deep learning applications in agriculture, highlighting the potential of neural networks for various agricultural tasks while identifying computational efficiency as a persistent challenge. Their subsequent work \cite{kamilaris_review_nodate} offered a focused analysis of convolutional neural networks in agriculture, demonstrating the growing importance of deep learning in agricultural applications. Recent research by Barbedo \cite{arnal_barbedo_plant_2019} explored the application of deep learning for plant disease identification, emphasizing the importance of developing efficient models for practical deployment in agricultural systems.\\
The specific challenges of deploying deep learning models in agricultural field conditions have been addressed by several studies. Mohanty et al. \cite{mohanty_frontiers_nodate} developed deep learning approaches for plant disease detection using smartphone images, highlighting the importance of model efficiency for mobile deployment. Fuentes et al. \cite{fuentes_robust_2017} proposed robust deep learning architectures for tomato anomaly detection under varying environmental conditions. Geetharamani and Arun Pandian \cite{g_identification_2019} investigated the application of convolutional neural networks for plant leaf disease detection, emphasizing the need for lightweight models suitable for resource-constrained environments. These studies established the foundation for intelligent agricultural monitoring systems while identifying the persistent challenge of balancing accuracy with computational constraints in real-world agricultural applications.\\
The development of efficient neural network architectures has been driven by the increasing demand for mobile and embedded applications. Howard et al. \cite{howard_mobilenets_2017} introduced MobileNets, which utilize depthwise separable convolutions to significantly reduce computational complexity while maintaining competitive accuracy. Sandler et al. \cite{sandler_mobilenetv2_2018} extended this work with MobileNetV2, incorporating inverted residuals and linear bottlenecks to further improve efficiency. Tan and Le \cite{tan_efficientnet_2020} proposed EfficientNet, demonstrating systematic scaling of network dimensions to achieve optimal accuracy-efficiency trade-offs. These architectural innovations have enabled the deployment of deep learning models on resource-constrained devices, though agricultural applications often require further optimization to meet stringent real-time and energy constraints.\\
Recent advances in knowledge distillation have focused on improving the transfer of knowledge between teacher and student networks. Romero et al. \cite{romero_fitnets_2015} introduced FitNets, which align intermediate representations between teacher and student models through hint-based learning. Park et al. \cite{park_relational_2019} proposed relational knowledge distillation, emphasizing the importance of preserving structural relationships in the transferred knowledge. Zagoruyko and Komodakis \cite{zagoruyko_paying_2017} developed attention transfer mechanisms that focus on important spatial locations during the distillation process. Tian et al. \cite{tian_contrastive_2022} introduced contrastive representation distillation, demonstrating improved knowledge transfer through contrastive learning principles. Abid Hussain et al \cite{hussain_gradual_2024} proposed a progressive knowledge transfer approach that systematically distilled hash code knowledge through incremental learning stages for improved large-scale image retrieval performance. Chen et al. \cite{chen_learning_2017} showed how knowledge distillation can improve object detection models, demonstrating the broader applicability of these techniques beyond classification tasks.\\
Traditional knowledge distillation approaches typically employ fixed temperature parameters throughout the training process \cite{hinton_distilling_2015}. However, recent research has explored dynamic temperature adjustment strategies to optimize knowledge transfer. Guo et al. \cite{guo_calibration_2017} investigated the effects of temperature scaling on model calibration, demonstrating that proper calibration can significantly improve model performance and reliability. This work highlighted the potential for adaptive temperature mechanisms to enhance knowledge distillation effectiveness.\\
The integration of advanced data augmentation techniques with knowledge distillation has shown promising results in enhancing model generalization. Zhang et al. \cite{zhang_mixup_2018} introduced Mixup, a data augmentation strategy that creates virtual training examples through convex combinations of input samples and their labels. Yun et al. \cite{yun_cutmix_2019} proposed CutMix, which combines patches from different images while mixing labels proportionally to the patch areas. DeVries and Taylor \cite{devries_improved_2017} developed improved regularization through cutout, while Zhong et al. \cite{zhong_random_2020} introduced random erasing for data augmentation. These techniques have proven particularly effective in improving model robustness and reducing overfitting, making them valuable components for knowledge distillation frameworks operating in diverse environmental conditions.\\
Despite significant progress in knowledge distillation and agricultural computer vision, existing approaches have not adequately addressed the specific requirements of agricultural embedded systems, where environmental variability, computational constraints, and real-time processing demands create unique challenges \cite{sharma_machine_2021}. Most knowledge distillation frameworks employ fixed temperature parameters that may not optimize knowledge transfer for different student model capacities \cite{yang_knowledge_2018}. Additionally, conventional data augmentation strategies may not capture the specific variations encountered in agricultural field conditions, such as varying lighting, weather conditions, and background complexity \cite{zheng_cropdeep_2019}.\\
The fundamental research problem lies in the trade-off between model accuracy and computational efficiency for agricultural embedded systems. Existing knowledge distillation methods fail to dynamically adapt to different student model capacities, resulting in suboptimal knowledge transfer and reduced performance on resource-constrained devices. Furthermore, standard augmentation techniques do not adequately prepare models for the diverse and challenging conditions encountered in agricultural environments, where factors such as natural lighting variations, complex backgrounds, and weather-dependent image quality significantly impact model performance.\\
This study addresses these limitations by proposing ATMS-KD (Adaptive Temperature and Mixed-Sample Knowledge Distillation), a novel framework specifically designed for agricultural embedded systems. The framework integrates capacity-aware adaptive temperature scheduling with mixed-sample augmentation techniques to optimize knowledge transfer from a MobileNetV3 Large teacher model to lightweight residual CNN students. The key contributions include: (1) development of an adaptive temperature scheduling mechanism that automatically adjusts distillation intensity based on student model capacity and performance gaps, (2) integration of complementary mixed-sample augmentation strategies to enhance model robustness for agricultural conditions, and (3) comprehensive evaluation demonstrating superior performance compared to existing knowledge distillation methods while maintaining computational efficiency suitable for embedded agricultural systems. The approach is evaluated on Rosa damascena maturity classification, demonstrating its effectiveness in achieving high accuracy while maintaining computational efficiency suitable for resource-constrained agricultural environments.

\section{Materials and Methods}
\label{sec:methods}

\subsection{Dataset and Preprocessing}
\label{sec:dataset}
\begin{figure*}[h!]
    \centering
    \begin{tabular}{cccc}
        \includegraphics[width=0.23\textwidth, height=0.23\textwidth]{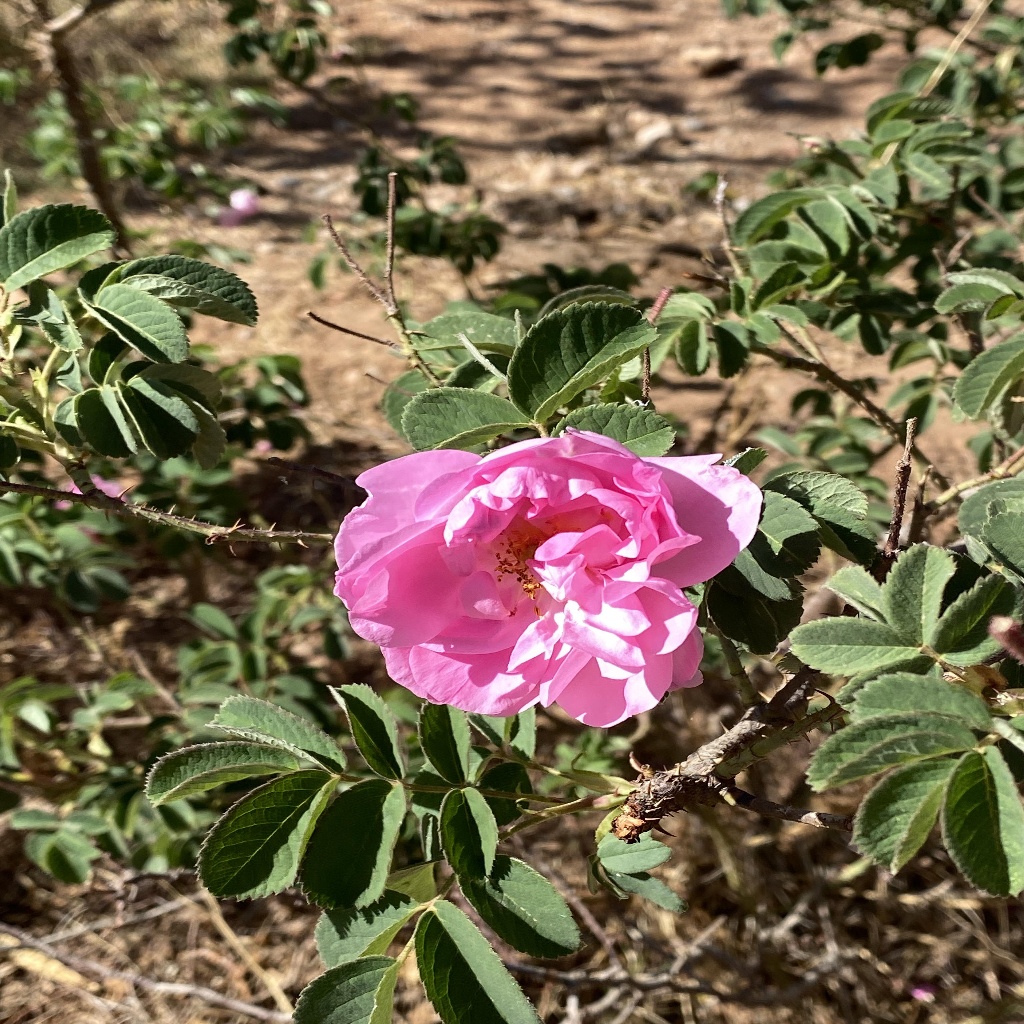} &
        \includegraphics[width=0.23\textwidth, height=0.23\textwidth]{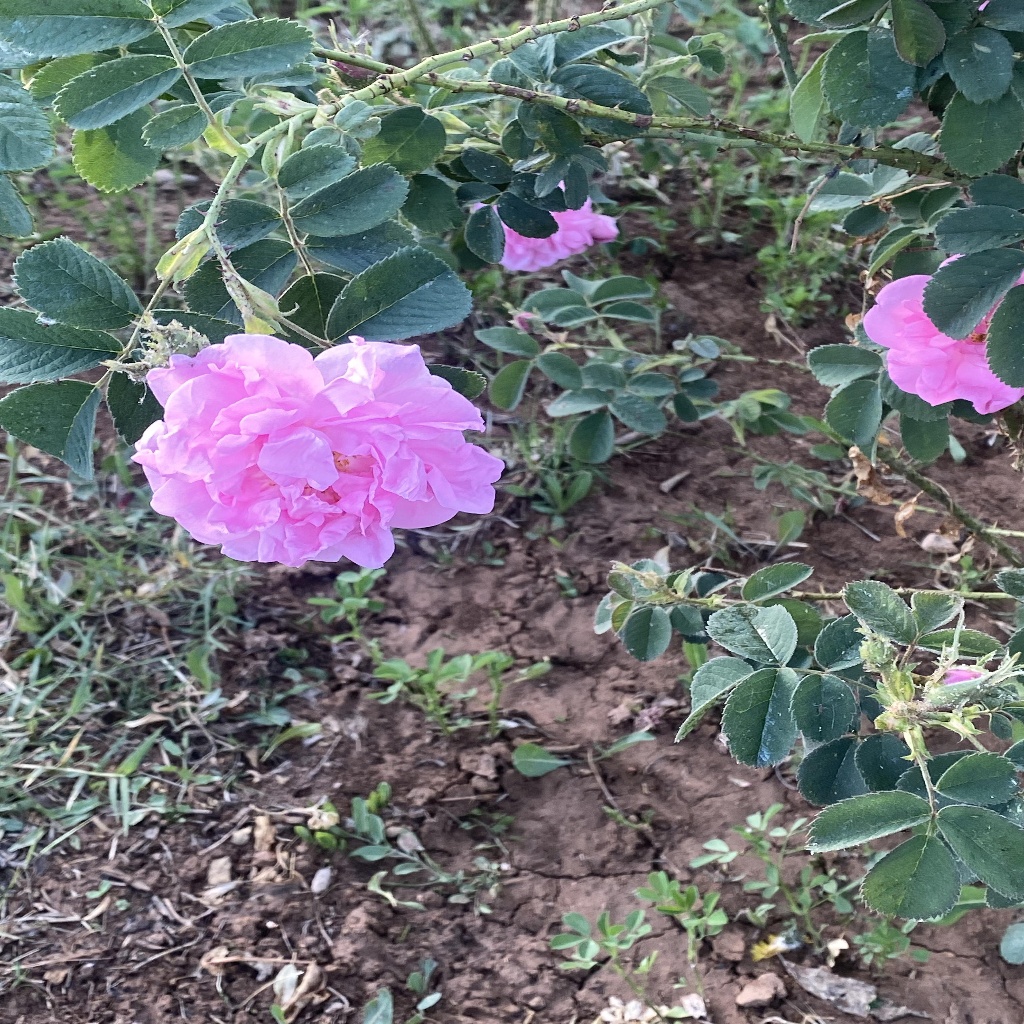} &
        \includegraphics[width=0.23\textwidth, height=0.23\textwidth]{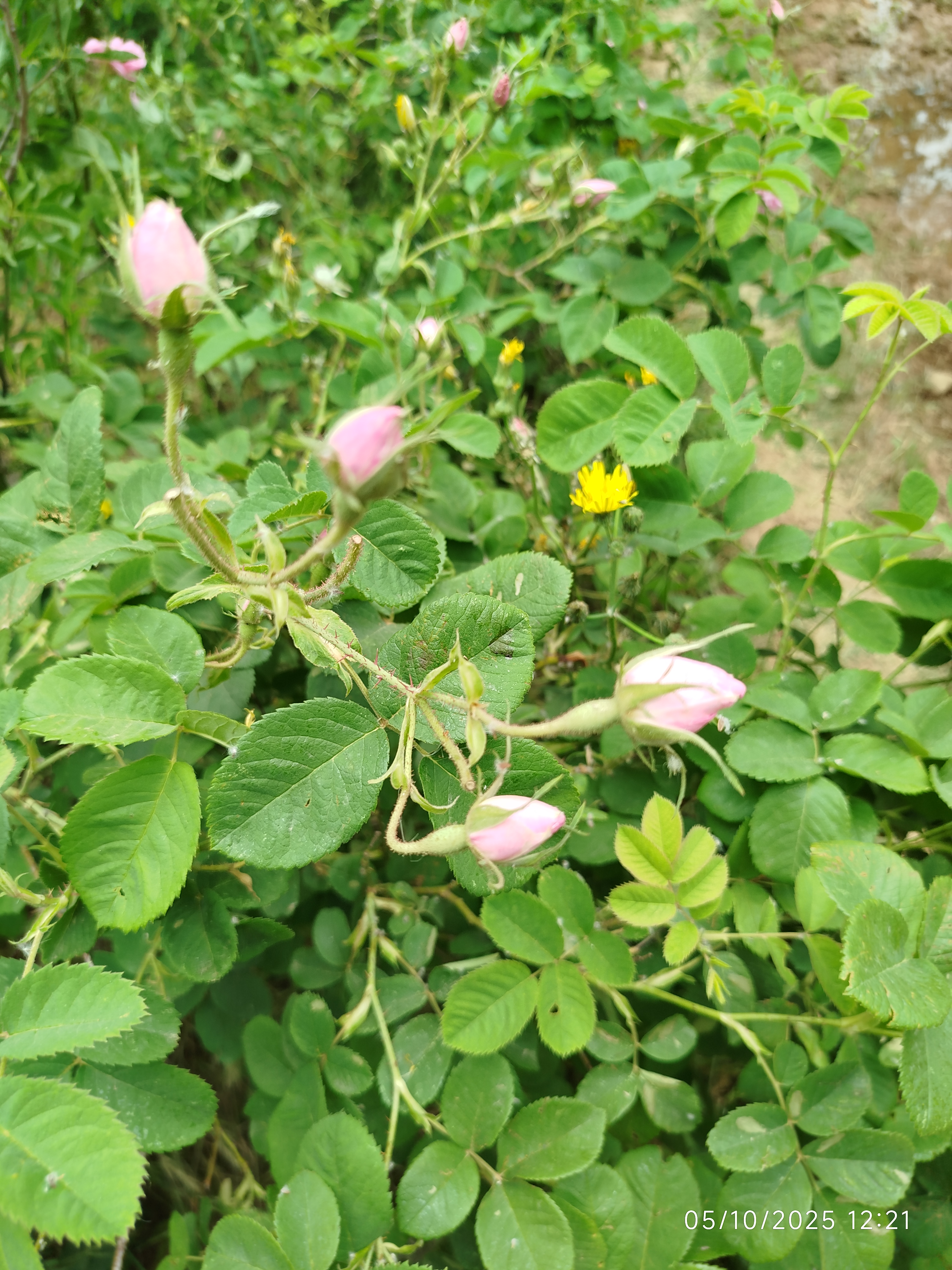} &
        \includegraphics[width=0.23\textwidth, height=0.23\textwidth]{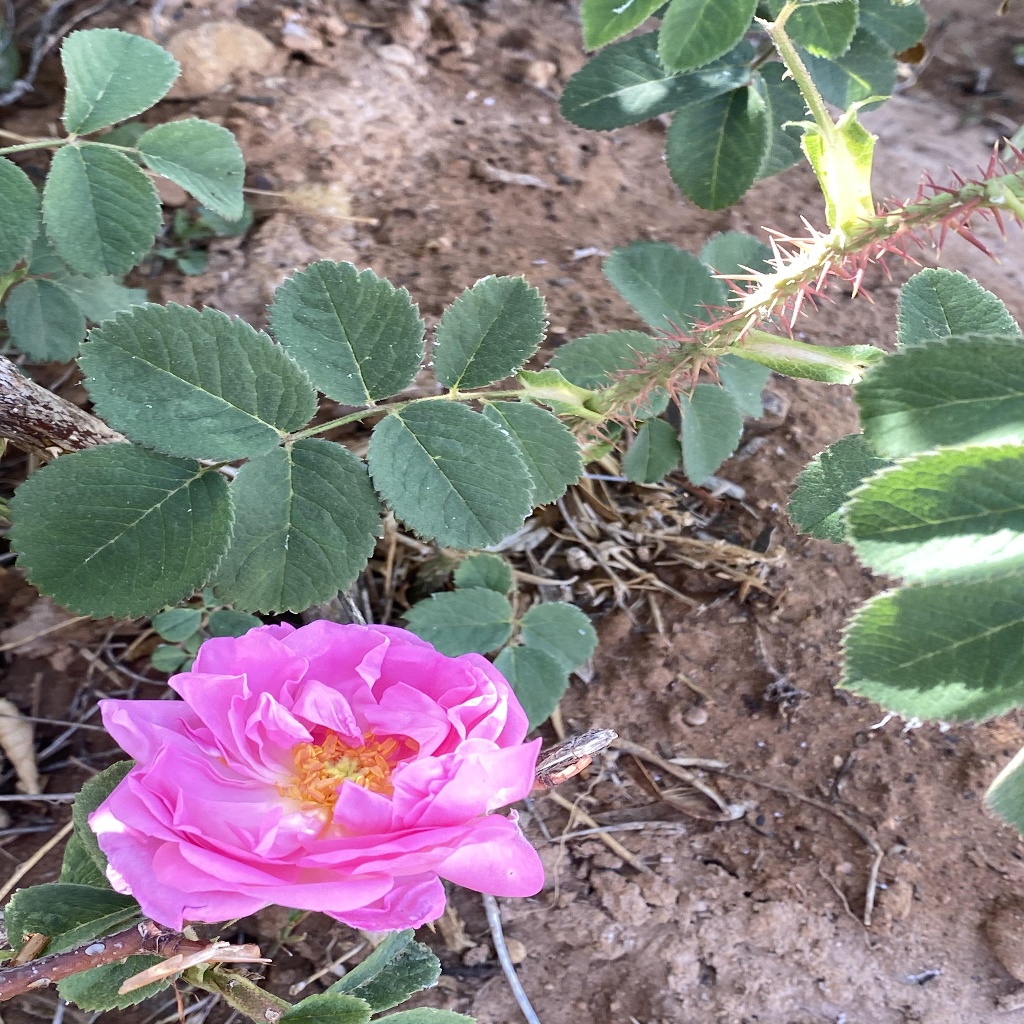} \\
        (a) & (b) & (c) & (d) \\

        \includegraphics[width=0.23\textwidth, height=0.23\textwidth]{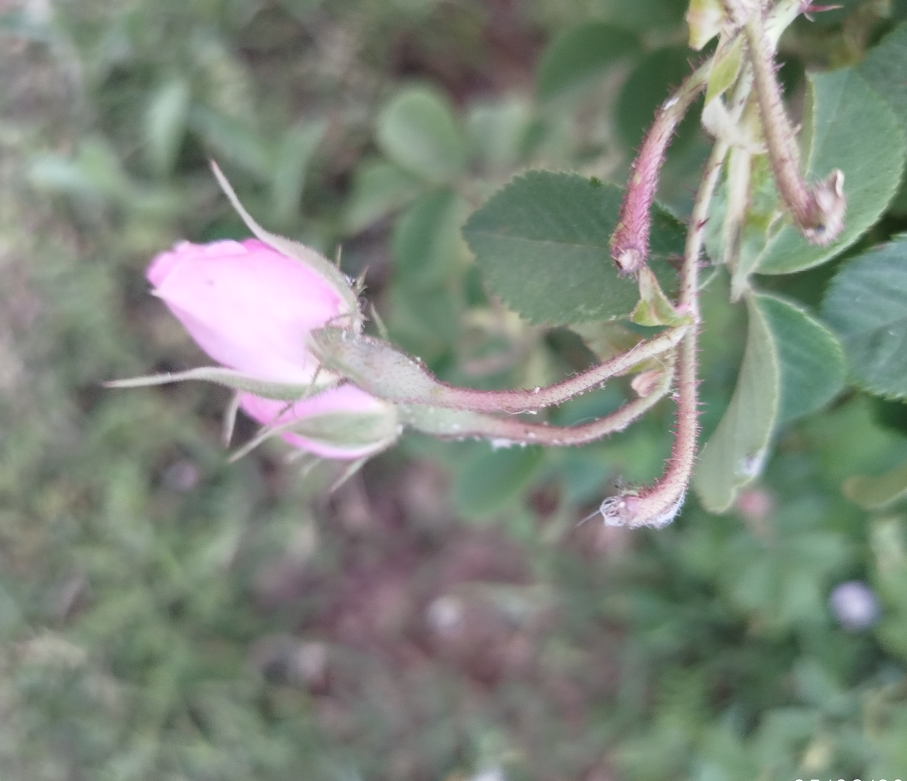} &
        \includegraphics[width=0.23\textwidth, height=0.23\textwidth]{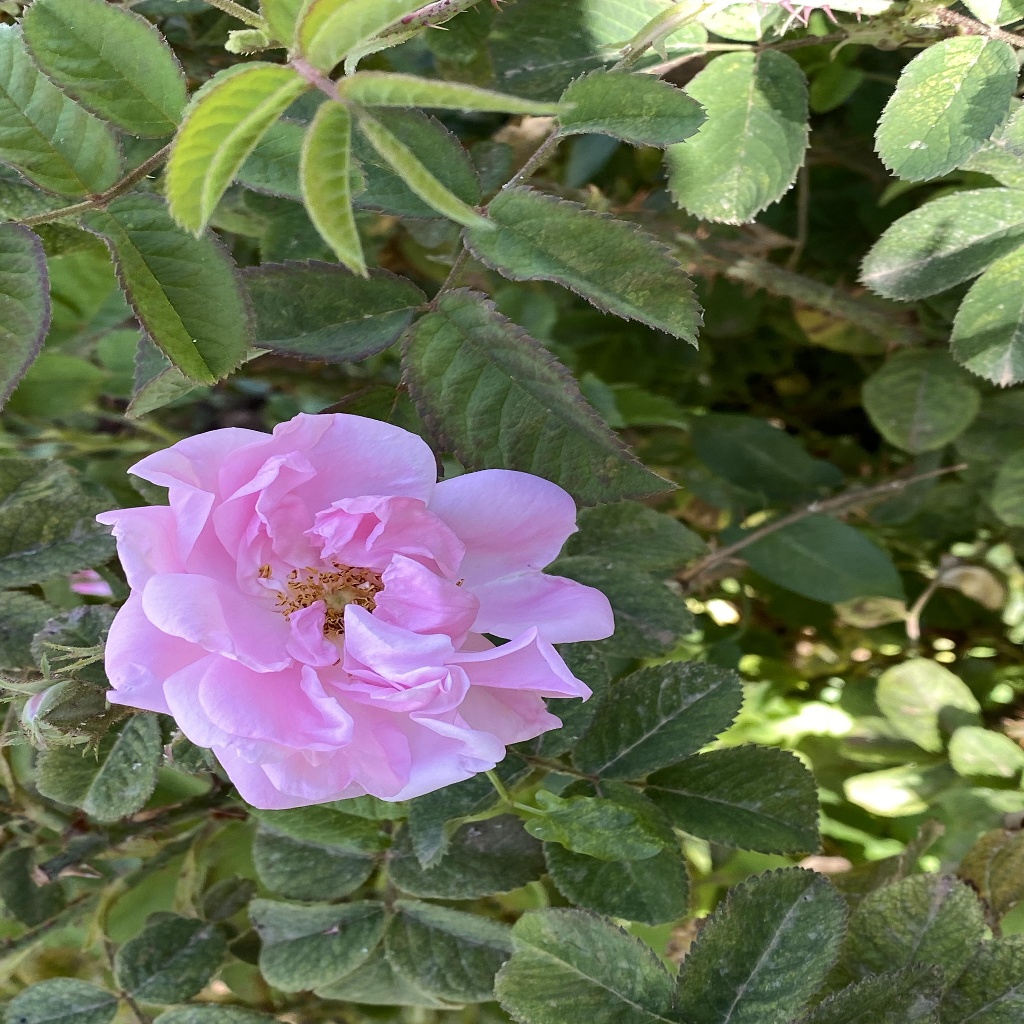} &
        \includegraphics[width=0.23\textwidth, height=0.23\textwidth]{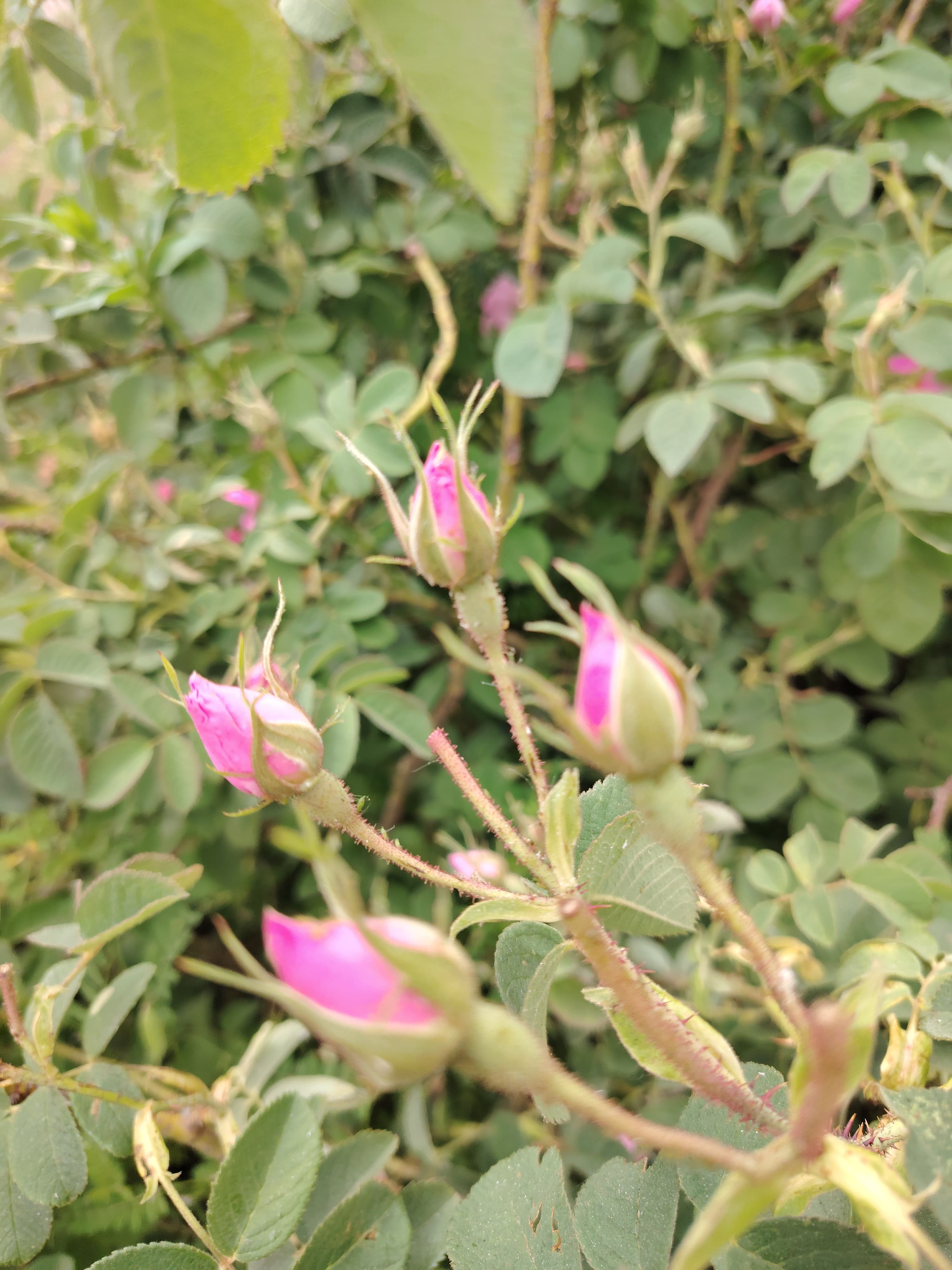} &
        \includegraphics[width=0.23\textwidth, height=0.23\textwidth]{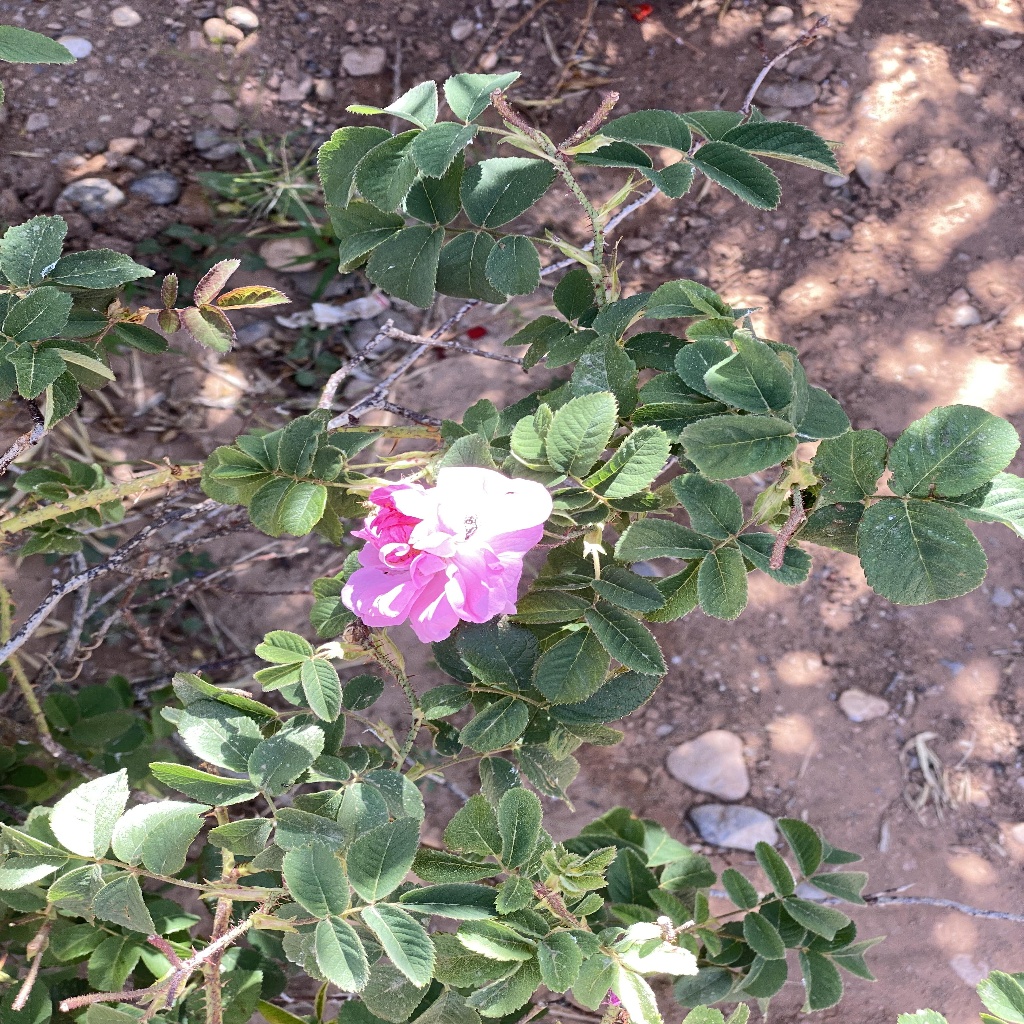} \\
        (e) & (f) & (g) & (h) \\
    \end{tabular}
    \caption{Examples of partial data sets. Note: (a) Sunny day, (b) Cloudy, (c) Long range, (d) Morning, (e) Evening, (f) Simple scenario, (g) High-density scenario, (h) Complicated background.}
    \label{fig:partial_data_sets}
\end{figure*}

The dataset used in this study consists of images of Rosa damascena (Damask rose) collected from agricultural fields in the Dades Oasis, southeastern Morocco  (Coordinates : 31.32\textdegree N, 5.33\textdegree W). Data acquisition was carried out during the flowering season using a smartphone at varying times of the day (morning, midday, evening) and under different weather conditions (sunny, cloudy), distances, and background complexities, capturing two maturity stages: immature and mature flowers (\textbf{Figure \ref{fig:partial_data_sets}}). \\
A total of 3114 images were collected and subsequently annotated into two categories: immature and mature. To maintain a balanced dataset, classes were sampled evenly whenever possible.

The dataset was divided into training and testing subsets. The training set was used for model learning, while the testing set was reserved for final performance evaluation. No overlap between the two subsets was allowed to avoid data leakage.

To enhance generalization, the dataset was pre-processed by resizing all images to a resolution of 224 × 224 pixels and normalizing pixel values using the mean and standard deviation of the ImageNet dataset. The diversity of capture conditions and backgrounds in this dataset provides a realistic benchmark for evaluating agricultural vision systems in uncontrolled environments.

\subsection{Model Architecture}
\subsubsection{Teacher model}
\begin{figure}[t]
  \centering
  \makebox[0pt]{\includegraphics[width=20cm]{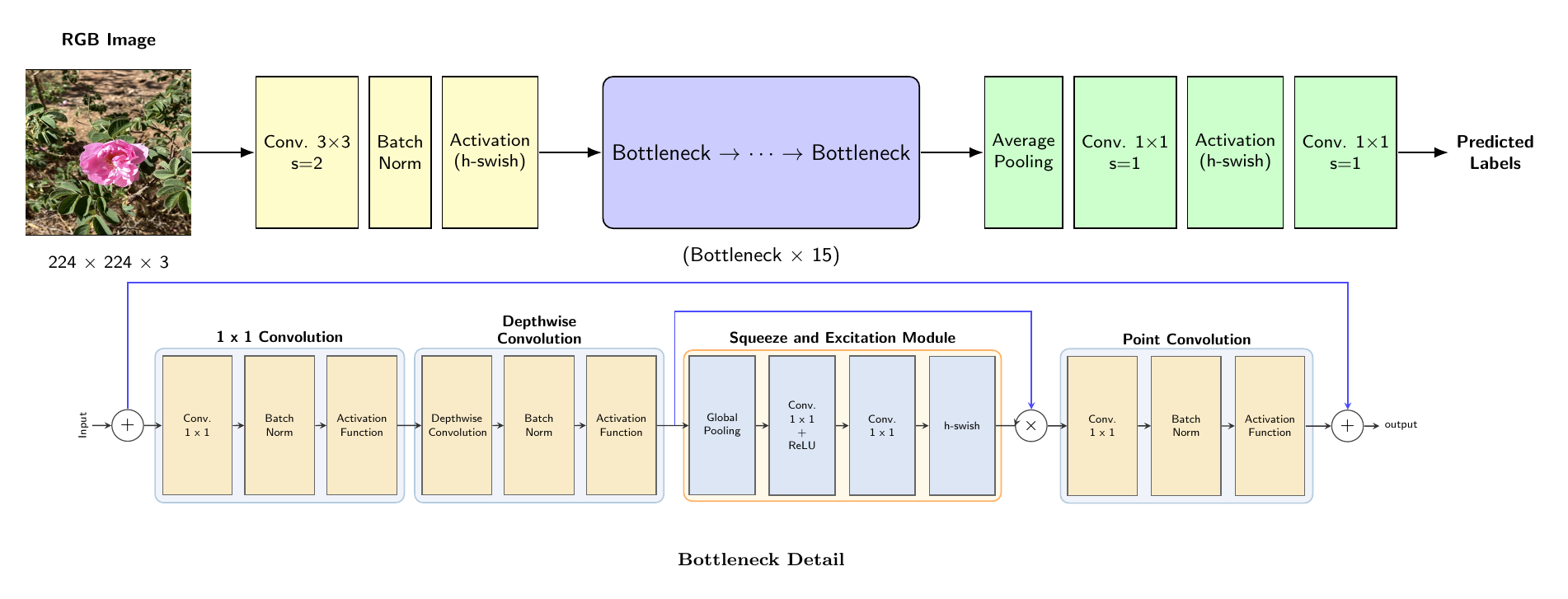}}
  \caption{Teacher model architecture: MobileNetV3 Large fine-tuned for \textit{Rosa damascena} maturity classification.}
  \label{fig:teacher-arch}
\end{figure}

The proposed framework follows a teacher–student knowledge distillation approach designed for efficient deployment on embedded agricultural systems. The teacher model is based on MobileNetV3 Large, a  convolutional neural network pretrained on the ImageNet dataset. The network was fine-tuned on the Rosa damascena dataset to adapt its feature extraction capabilities to the target classification task. The final classifier layer of the teacher model was replaced with a custom fully connected structure to match the two output classes (immature and mature).\\
MobileNetV3 Large architecture (\textbf{Figure \ref{fig:teacher-arch}}) begins with a \(3 \times 3\) convolution 
and activation (hard-swish), followed by a sequence of 15 bottleneck blocks. 
Each bottleneck block consists of:\\
- Pointwise convolution for channel expansion, a depthwise convolution for spatial filtering, a Squeeze-and-Excitation (SE) module for adaptive channel reweighting, and a final \(1 \times 1\) projection convolution to reduce dimensionality.\\
- Residual connections are applied when input and output dimensions match, enabling efficient gradient flow. The network concludes with a global average pooling  layer and a \(1 \times 1\) convolution before the classification layer. This design allows MobileNetV3 Large to achieve high accuracy while 
keeping the number of parameters and operations low, making it well-suited  as a teacher network in knowledge distillation.\\ 

\subsubsection{Student model}
\label{sec:student_model}
\begin{figure}[htbp]
    \centering
    \includegraphics[width=0.75\textwidth]{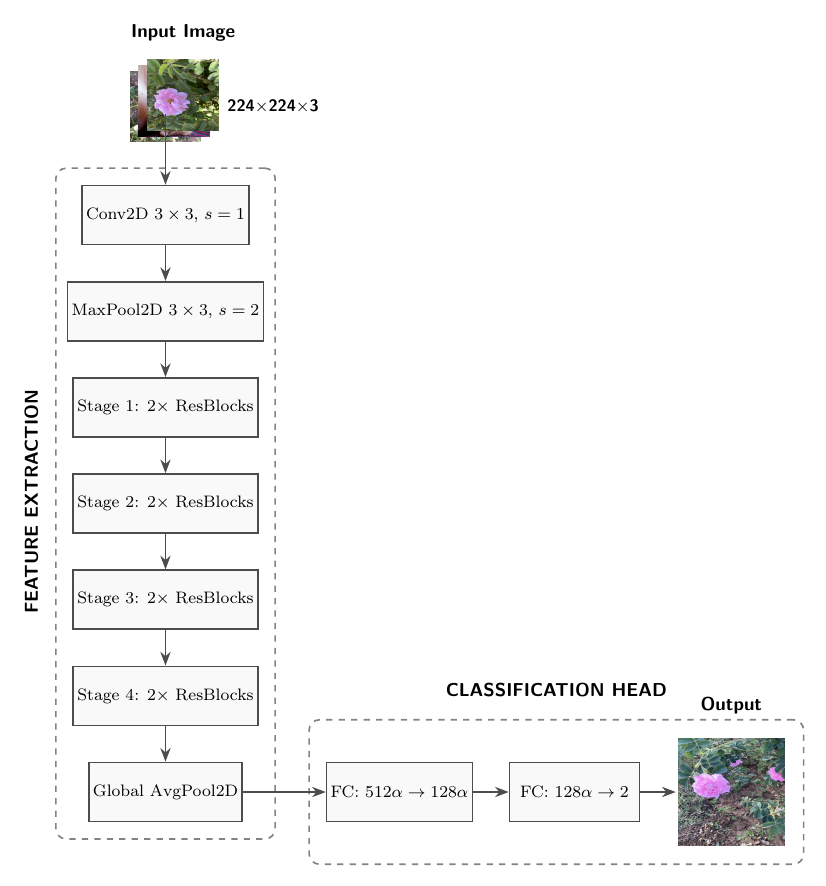}
    \caption{Overall architecture of the proposed student network based on the Lightweight Residual CNN.}
    \label{fig:custom_cnn}
\end{figure}

We developed a student model based on a lightweight residual convolutional neural network (CNN) (\textbf{Figure \ref{fig:custom_cnn}})
designed for fast inference on resource-limited devices. 
The network starts with a \(3 \times 3\) convolutional layer followed by a max pooling layer, 
and then proceeds through four stages of residual blocks with gradually increasing channel sizes. 
Each residual block follows a bottleneck structure, combining \(1 \times 1\) and \(3 \times 3\) convolutions 
with batch normalization, dropout, and skip connections to improve training stability and efficiency. 
The model ends with an adaptive average pooling layer and a fully connected classifier.

\begin{figure}[htbp]
    \centering
    \includegraphics[width=0.85\textwidth]{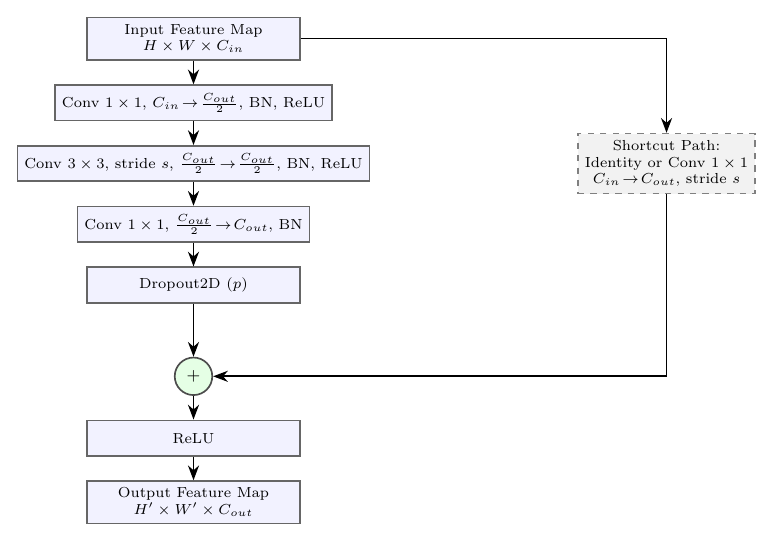}
    \caption{Architecture of the Lightweight Residual Block used in the proposed student model.}
    \label{fig:light_res_block}
\end{figure}

Each Lightweight Residual Block takes (\textbf{Figure \ref{fig:light_res_block}}) an input feature map of size \(H \times W \times C_{\mathrm{in}}\) and applies:
\begin{enumerate}
    \item A \(1 \times 1\) convolution to reduce channel dimensions from \(C_{\mathrm{in}}\) to \(C_{\mathrm{out}}/2\), followed by batch normalization and ReLU activation.
    \item A \(3 \times 3\) convolution with stride \(s\), keeping the channel size at \(C_{\mathrm{out}}/2\), followed by batch normalization and ReLU activation.
    \item A final \(1 \times 1\) convolution to expand channels from \(C_{\mathrm{out}}/2\) to \(C_{\mathrm{out}}\), followed by batch normalization and spatial dropout.
\end{enumerate}
A shortcut path, either identity mapping or a \(1 \times 1\) convolution (for dimension or stride changes), 
is added to the main branch output, and the result is passed through a ReLU activation to produce the block output.

To evaluate the balance between accuracy and computational cost, 
three versions of the student network were tested by adjusting the width multiplier 
(\(0.75\times\), \(1.0\times\), and \(1.25\times\)), which proportionally scales the number of channels in each layer. 
Both the teacher and student models take RGB images of size \(224 \times 224\) as input. 
During training, the teacher model transfers knowledge to the student using the proposed ATMS-KD method, 
while only the student model is deployed for real-time inference on the embedded platform.






\subsection{ATMS-KD Framework}
The Adaptive Temperature and Mixed Sample Knowledge Distillation (ATMS-KD) framework combines two effective strategies-\textit{Mixed Sample Augmentation} and \textit{Adaptive Temperature Scheduling}—to improve the training of a lightweight student model via knowledge transfer from a larger teacher network.

\subsubsection{Mixup }
Mixup is a data augmentation technique introduced by Zhang et al. in ICLR 2018~\cite{zhang_mixup_2018} that trains neural networks using convex combinations of pairs of training samples and their labels. Given two example-label pairs $(x_i, y_i)$ and $(x_j, y_j)$, the method generates a synthetic training sample as:

\begin{equation}
\tilde{x} = \lambda x_i + (1 - \lambda) x_j, \quad 
\tilde{y} = \lambda y_i + (1 - \lambda) y_j,
\end{equation}

\noindent where $\lambda \sim \text{Beta}(\alpha, \alpha)$ is sampled from a Beta distribution. Mixup encourages models to exhibit linear behavior between examples, improving generalization, reducing overfitting, and enhancing robustness to adversarial examples. The technique has proven particularly effective in stabilizing training processes across various deep learning applications~\cite{zhang_mixup_2018}.

\subsubsection{CutMix}
CutMix was proposed by Yun et al. in ICCV 2019 as a region-based augmentation strategy~\cite{yun_cutmix_2019} that replaces a randomly selected rectangular patch in one image with a patch from another image. Unlike erasing methods that remove information, CutMix preserves spatial context while introducing variability. The augmentation process is formalized as:

\begin{equation}
\tilde{x} = \mathbf{M} \odot x_i + (1 - \mathbf{M}) \odot x_j, \quad 
\tilde{y} = \lambda y_i + (1 - \lambda) y_j,
\end{equation}

\noindent where $\mathbf{M}$ is a binary mask indicating the patch location, and $\lambda$ corresponds to the area ratio of the pasted patch. This technique retains more spatial context than erasing methods and has consistently outperformed prior approaches on classification and localization tasks, including CIFAR, ImageNet, and object detection benchmarks. Additionally, CutMix improves robustness to input corruption and out-of-distribution detection~\cite{yun_cutmix_2019}.
    
\subsubsection{Complementary Effects of Mixup and CutMix}

\noindent Mixup and CutMix provide complementary benefits for deep learning applications. Mixup improves robustness through global linear interpolation between entire images, while CutMix enhances local feature learning by preserving spatial structures and introducing region-level variations. Recent theoretical work shows that mixed-sample data augmentation techniques, including both Mixup and CutMix, act as forms of input gradient regularization, enhancing both generalization and adversarial robustness. The integration of both strategies in the ATMS-KD framework enriches the training signal, which proves particularly valuable for agricultural datasets characterized by variations in lighting conditions, background complexity, and object density. This combination ensures comprehensive data augmentation that addresses both global and local feature learning requirements for robust agricultural image classification.

\subsubsection{Adaptive Temperature Scheduling}

Traditional knowledge distillation uses a fixed temperature \(T\) to soften teacher outputs for student learning. However, the optimal \(T\) can vary depending on the student's capacity, the learning stage, and the performance gap between teacher and student. Research indicates that dynamic temperature adjustment may yield better results than a fixed value.\\

\begin{algorithm}[H]
\caption{Temperature Sensitivity Analysis}
\label{alg:temp_sensitivity}
\begin{algorithmic}[1]
\Require Logits $z \in \mathbb{R}^C$, temperature range $[\tau_{\min}, \tau_{\max}]$, validation set $\mathcal{D}_{\text{val}}$
\Ensure Optimal temperature $\tau^\star$
\State
\For{$\tau \in [\tau_{\min}, \tau_{\max}]$}
    \State $p(\tau) \gets \text{Softmax}(z / \tau)$ \Comment{Temperature-scaled probabilities}
    \State $H(\tau) \gets -\sum_{i=1}^C p_i(\tau) \log p_i(\tau)$ \Comment{Entropy}
    \State $\text{Acc}(\tau) \gets \text{EvaluateTemp}(S, \mathcal{D}_{\text{val}}, \tau)$
\EndFor
\State
\State $\tau^\star \gets \arg\max_{\tau} \left[ H(\tau) \cdot \text{Acc}(\tau) \right]$
\State
\State \Return $\tau^\star$
\end{algorithmic}
\end{algorithm}

In ATMS-KD, the Adaptive Temperature Scheduler adjusts \(T\) based on (Algorithm \ref{alg:temp_sensitivity}):

- Student capacity: Smaller student models use a higher initial \(T\) to capture richer soft-label information.\\
 
- Performance gap: A larger teacher-student accuracy gap raises \(T\), emphasizing soft supervision.\\

- Training progress: As training advances, \(T\) gradually decreases, encouraging focus on true hard labels.
This adaptive schedule helps the student benefit from informative soft targets in early training while refining its learning toward hard labels as it converges.Traditional knowledge distillation approaches employ a fixed temperature parameter \(T\) to soften teacher model outputs for student learning \cite{hinton_distilling_2015}. However, the optimal temperature value can vary significantly depending on multiple factors, including the student model's capacity, the current learning stage, and the performance gap between teacher and student networks. Recent research indicates that dynamic temperature adjustment strategies may yield substantially better results than fixed-value approaches \cite{guo_calibration_2017}.
In ATMS-KD, the Adaptive Temperature Scheduler dynamically adjusts the temperature parameter \(T\)  throughout the training process based on three key principles (Algorithm \ref{alg:temp_sensitivity}):\\
Student Capacity Adaptation: Smaller student models require higher initial temperature values to effectively capture richer soft-label information from the teacher network. This approach compensates for the reduced representational capacity of compact architectures by increasing the informativeness of the soft targets during knowledge transfer.\\
Performance Gap Compensation: When a larger accuracy gap exists between teacher and student models, the scheduler increases the temperature value to emphasize soft supervision over hard label learning. This mechanism ensures that students with significant performance deficits receive more guidance from the teacher's probability distributions.\\ 
Training Progress Awareness: As training progresses and the student model's performance improves, the temperature gradually decreases to encourage increased focus on the true hard labels. This progressive refinement helps the student transition from relying heavily on teacher guidance to developing independent decision-making capabilities.\\
This adaptive scheduling strategy enables the student model to benefit from informative soft targets during early training phases while progressively refining its learning toward hard labels as convergence approaches. The dynamic adjustment mechanism addresses the limitations of fixed-temperature approaches by automatically optimizing the knowledge transfer intensity based on the student's evolving capacity and performance characteristics throughout the training process.

\begin{figure}[t]
    \centering
    \includegraphics[width=1\textwidth]{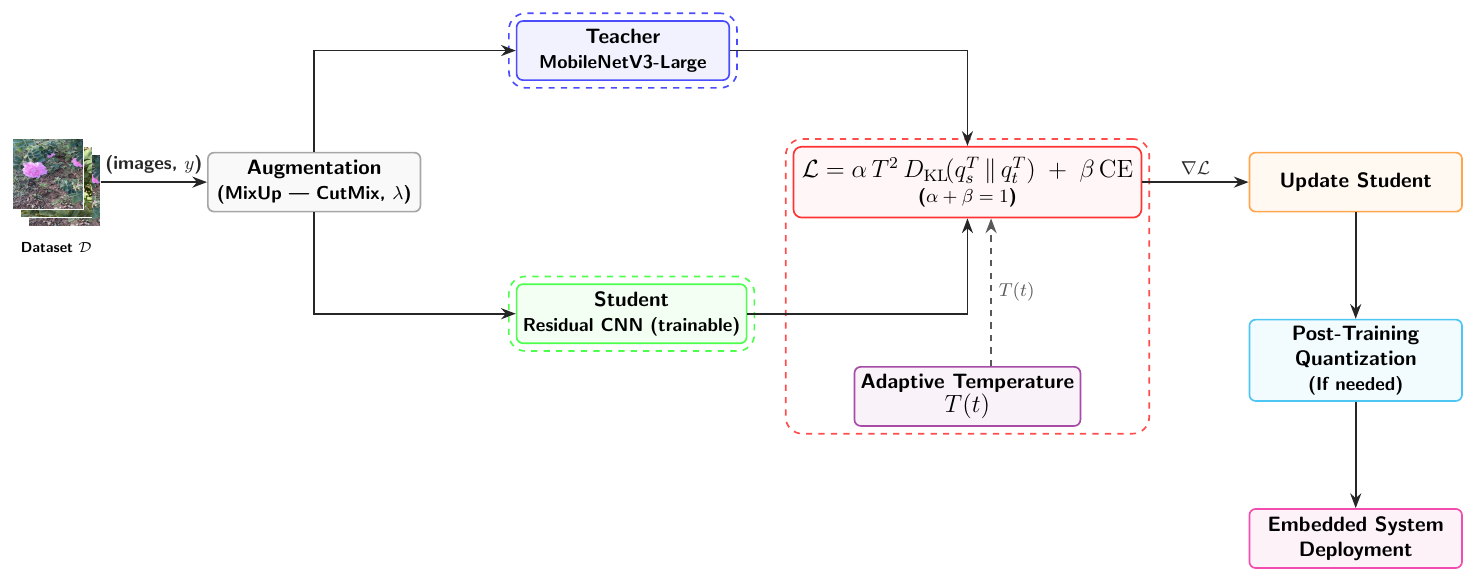}
    \caption{Overview of the ATMS-KD framework: combining adaptive temperature scheduling with mixed-sample augmentation techniques.}
  \label{fig:atmskd}
\end{figure}

The complete ATMS-KD training pipeline proceeds as follows (\textbf{Figure~\ref{fig:atmskd}} and \textbf{Algorithm~\ref{alg:kd_training}}):

\begin{algorithm}[H]
\caption{Knowledge Distillation Training Framework}
\label{alg:kd_training}
\begin{algorithmic}[1]
\Require Training dataset $\mathcal{D} = \{(x_i, y_i)\}_{i=1}^N$, pre-trained teacher $T$, student $S$, temperature $\tau > 0$, weights $\alpha, \beta \in [0,1]$ with $\alpha+\beta=1$, total epochs $E$
\Ensure Trained student model $S^\star$
\State
\State \textbf{Teacher Preparation:}
\If{$T$ is not pre-trained}
    \State Train $T$ on $\mathcal{D}$ until convergence
    \State Save best weights as $T^\star$
\EndIf
\State Freeze all parameters of $T$
\State
\State \textbf{Student Initialization:}
\State Randomly initialize $S$
\State Set optimizer $\text{opt}_S$
\State
\For{$\text{epoch} = 1$ \textbf{to} $E$}
    \For{\textbf{each} mini-batch $(x, y) \subset \mathcal{D}$}
        \State \Comment{Forward pass}
        \State $z_T \gets T(x)$ \Comment{Teacher logits (no gradient)}
        \State $z_S \gets S(x)$ \Comment{Student logits}
        \State
        \State \Comment{Soft targets from teacher}
        \State $p_T \gets \text{Softmax}(z_T / \tau)$
        \State $p_S \gets \text{LogSoftmax}(z_S / \tau)$
        \State
        \State \Comment{Loss computation}
        \State $\mathcal{L}_{\text{KD}} \gets \tau^2 \cdot \text{KL}(p_S, p_T)$
        \State $\mathcal{L}_{\text{CE}} \gets \text{CrossEntropy}(z_S, y)$
        \State $\mathcal{L}_{\text{total}} \gets \alpha \mathcal{L}_{\text{KD}} + \beta \mathcal{L}_{\text{CE}}$
        \State
        \State \Comment{Backpropagation}
        \State $\text{opt}_S.\text{zero\_grad}()$
        \State $\mathcal{L}_{\text{total}}.\text{backward}()$
        \State $\text{opt}_S.\text{step}()$
    \EndFor
    \State
    \State \Comment{Validation and checkpoint}
    \State $\text{acc}_{\text{val}} \gets \text{Evaluate}(S, \mathcal{D}_{\text{val}})$
    \If{$\text{acc}_{\text{val}} > \text{best\_accuracy}$}
        \State $\text{best\_accuracy} \gets \text{acc}_{\text{val}}$
        \State $S^\star \gets S$
    \EndIf
\EndFor
\State
\State \Return $S^\star$
\end{algorithmic}
\end{algorithm}

\section{Evaluation Metrics}
The performance of the proposed ATMS-KD framework was evaluated using multiple metrics to provide comprehensive assessment across different aspects of model performance. This section presents the mathematical formulations and definitions of all evaluation metrics used in the comparative analysis.

\subsection{Classification Metrics}
Accuracy measures the proportion of correctly classified samples over the total number of samples:
\begin{equation}
\text{Accuracy} = \frac{TP + TN}{TP + TN + FP + FN} \times 100\%
\end{equation}
where $TP$, $TN$, $FP$, and $FN$ represent true positives, true negatives, false positives, and false negatives, respectively.

\noindent Precision quantifies the proportion of true positive predictions among all positive predictions:
\begin{equation}
\text{Precision} = \frac{TP}{TP + FP} \times 100\%
\end{equation}

\noindent Recall (also known as sensitivity) measures the proportion of true positive cases that were correctly identified:
\begin{equation}
\text{Recall} = \frac{TP}{TP + FN} \times 100\%
\end{equation}

\noindent F1-Score provides the harmonic mean of precision and recall, offering a balanced measure when dealing with imbalanced datasets:
\begin{equation}
\text{F1-Score} = 2 \times \frac{\text{Precision} \times \text{Recall}}{\text{Precision} + \text{Recall}} \times 100\%
\end{equation}

\subsection{Knowledge Distillation Metrics}
Knowledge Retention (KR) evaluates how effectively the student model preserves the teacher's performance:
\begin{equation}
\text{KR} = \frac{\text{Student Accuracy}}{\text{Teacher Accuracy}} \times 100\%
\end{equation}
This metric provides insight into the effectiveness of the knowledge transfer process, with higher values indicating better preservation of teacher model capabilities.

\subsection{Computational Performance Metrics}
Average Inference Time measures the mean time required for model prediction on a single sample:
\begin{equation}
\text{Avg Inference Time} = \frac{1}{N} \sum_{i=1}^{N} t_i
\end{equation}
where $t_i$ represents the inference time for the $i$-th sample and $N$ is the total number of test samples.

\noindent Throughput quantifies the number of samples processed per second:
\begin{equation}
\text{Throughput} = \frac{1}{\text{Avg Inference Time}} \quad \text{(samples per second)}
\end{equation}

\noindent Model Size refers to the storage requirements of the trained model in megabytes (MB), including all parameters and network architecture information.

\noindent Parameter Count represents the total number of trainable parameters in the model, typically expressed in millions (M) for deep neural networks.

\subsection{Experimental Setup}
All experiments were conducted on a workstation equipped with an NVIDIA RTX 4060Ti GPU (16~GB VRAM), an AMD Ryzen~7 5700X 8-core processor, and 32~GB of RAM. The training environment used PyTorch~2.6 with CUDA~11.8 support on Ubuntu~22.04~LTS. All random seeds were fixed to 42 to ensure reproducibility across experimental runs.\\
The dataset described in Section~\ref{sec:dataset} was split into 70\% for training and 30\% for testing, ensuring that images from the same scene did not appear in multiple sets. All images were resized to $224 \times 224$ pixels and normalized using the mean and standard deviation of ImageNet.\\
The teacher model was MobileNetV3 Large, pre-trained on ImageNet and fine-tuned on the \textit{Rosa damascena} dataset. The student model was the custom Lightweight Residual CNN described in Section~\ref{sec:student_model}. Teacher training used the AdamW optimizer with an initial learning rate of $1 \times 10^{-3}$, while student training employed a learning rate of $2 \times 10^{-3}$. Both models utilized cosine annealing with warm restarts for learning rate scheduling. The batch size was set to 16 for teacher training and 32 for student training. Training lasted for 30 epochs for the teacher and 80 epochs for the student models.\\
For the proposed ATMS-KD framework, Mixup and CutMix augmentation strategies were applied with a probability of 0.5 each time augmentation was triggered. The adaptive temperature scheduler was capacity-aware, initializing at different values based on the student model width multipliers: compact models ($0.75\times$) started at $T = 6.0$, standard models ($1.0\times$) at $T = 4.5$, and enhanced models ($1.25\times$) at $T = 4.3$. The temperature gradually annealed to approximately $T = 3.0$ for all configurations as training progressed.\\
The knowledge distillation loss was composed of a weighted combination of distillation loss ($\alpha = 0.7$) and hard target loss ($\beta = 0.3$), along with L2 regularization ($\gamma = 1 \times 10^{-5}$).

\section{Results}
\subsection{Teacher Model Performance }

\begin{figure}[htbp]
    \centering
    \includegraphics[width=1\textwidth]{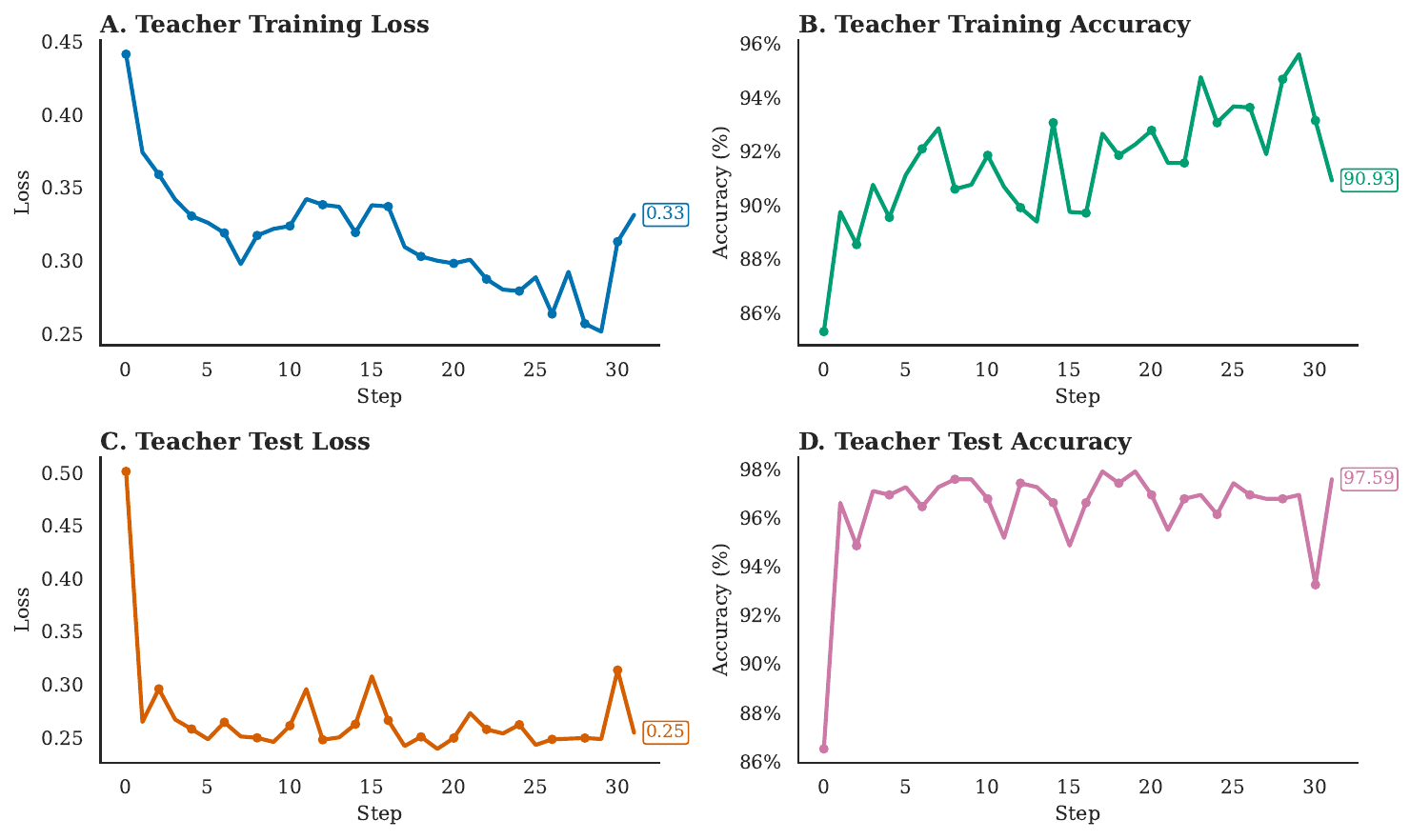}
    \caption{Performance metrics of the teacher model (MobileNetV3 Large).}
    \label{fig:teacher_metrics}
\end{figure}

The MobileNetV3 Large teacher model with 5.7M total parameters was fine-tuned on the Rosa damascena dataset and achieved strong performance, as shown in \textbf{Figure \ref{fig:teacher_metrics}}. The model converged within 15 epochs, with training loss decreasing from 0.45 to 0.33 and training accuracy reaching 90.93\%. The regularization techniques used, including label smoothing ($\varepsilon=0.1$) and weight decay ($\lambda=1\mathrm{e}{-4}$), helped maintain stable training without overfitting.\\
The model showed excellent generalization, achieving 97.59\% test accuracy with a test loss of 0.25, which was lower than the training loss of 0.33. This indicates that the model generalizes well to new Rosa damascena samples, making it suitable as a teacher for knowledge distillation. The superior performance of the teacher model establishes a strong foundation for effective knowledge transfer to student networks. However, the high computational requirements of MobileNetV3 Large make it unsuitable for deployment on resource-limited agricultural devices, which creates the need to transfer its knowledge to smaller student models that can operate efficiently in embedded agricultural systems.

\subsection{Student Model Architecture Evaluation
}

Three student variants—Compact, Standard, and Enhanced—were derived from the same residual CNN backbone by adjusting the width multipliers to 
$0.75\times$, 
$1.0\times$, and 
$1.25\times$, respectively. These modifications resulted in controlled differences in channel width per stage, enabling an empirical exploration of the trade-off between efficiency and accuracy.

\subsubsection{Compact Configuration Performance}
\begin{figure}[htbp]
    \centering
    \begin{subfigure}{0.5\textwidth}
        \centering
        \includegraphics[width=\linewidth]{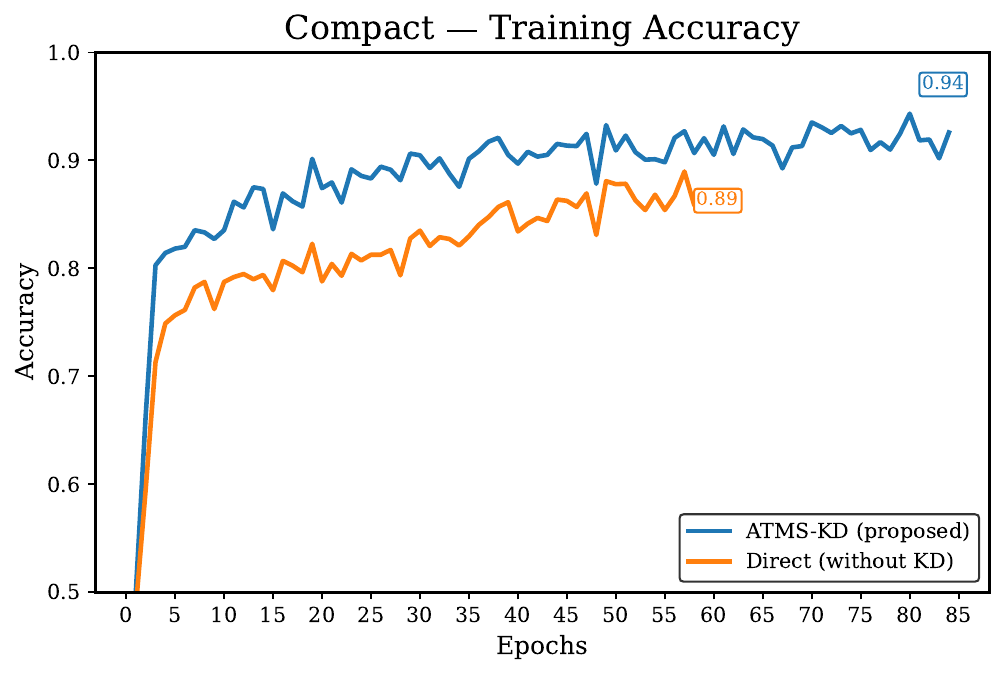}
        \caption{Compact Training Accuracy (KD vs Direct).}
        \label{fig:compact_train_acc}
    \end{subfigure}\hfill
    \begin{subfigure}{0.5\textwidth}
        \centering
        \includegraphics[width=\linewidth]{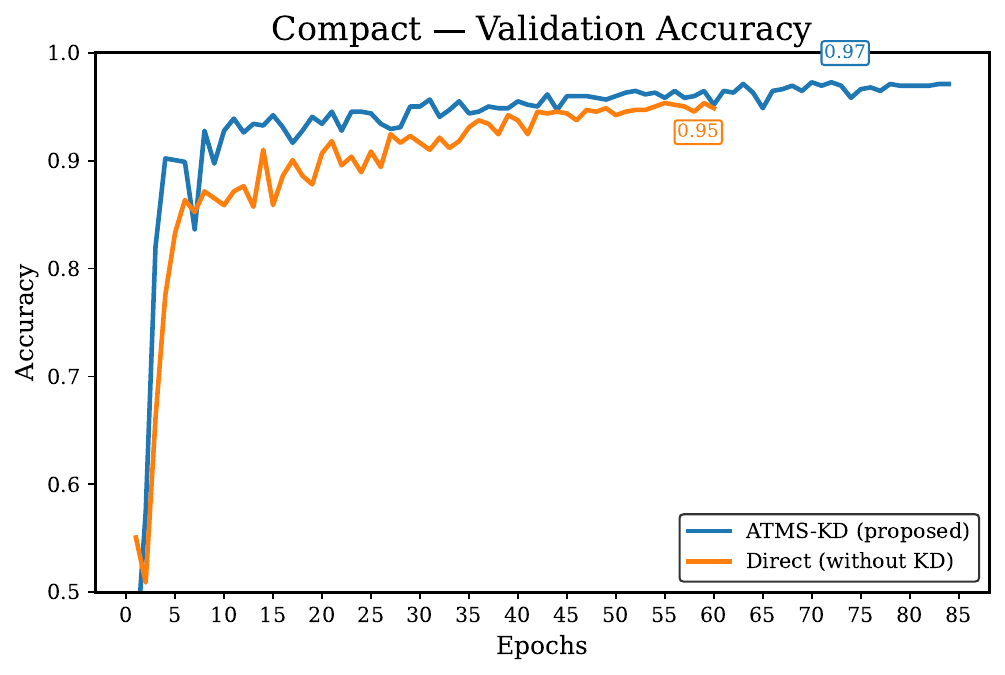}
        \caption{Compact Validation Accuracy (KD vs Direct).}
        \label{fig:compact_val_acc}
    \end{subfigure}
    
    \begin{subfigure}{0.5\textwidth}
        \centering
        \includegraphics[width=\linewidth]{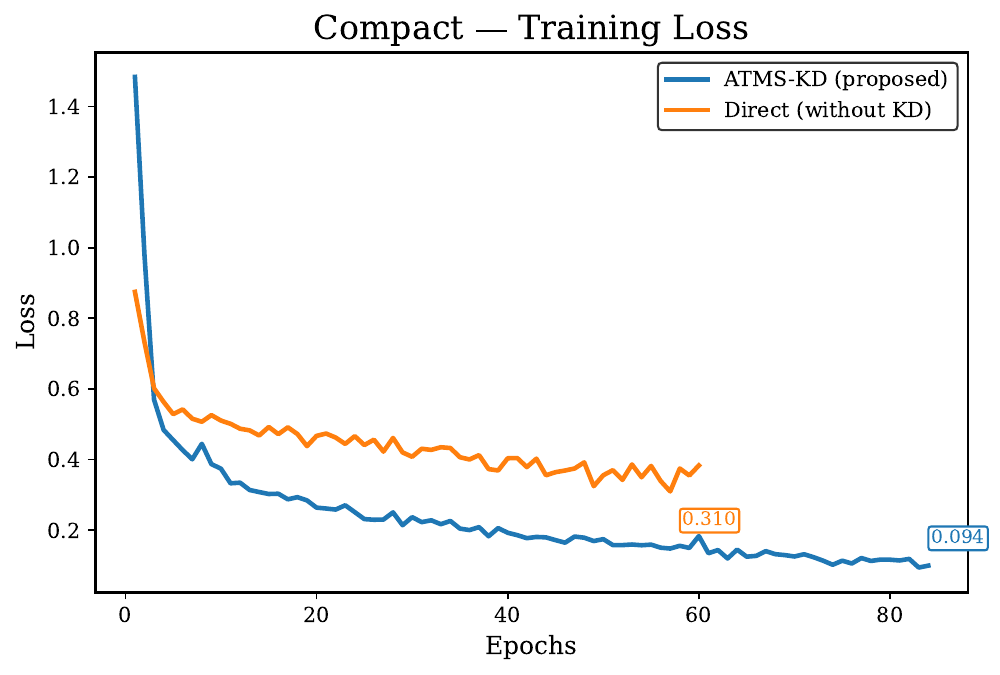}
        \caption{Compact Training Loss (KD vs Direct).}
        \label{fig:compact_loss}
    \end{subfigure}

    \caption{Training and validation performance of the Compact student model using ATMS-KD vs Direct training.}
    \label{fig:compact_metrics}
\end{figure}

The compact student model, designed for highly resource-constrained environments, contained 1.3M parameters and utilized a width multiplier of 0.75×. \textbf{Figure \ref{fig:compact_metrics}} presents the training and validation performance of this configuration across multiple epochs. The compact model achieved significant performance improvements through the proposed ATMS-KD approach, reaching 94\% training accuracy compared to 89\% with direct training methods—a 5 percentage point improvement. The validation accuracy demonstrated even greater gains, increasing from 95\% with direct training to 97\% with ATMS-KD, representing a 2 percentage point improvement and achieving the highest accuracy among all three model configurations at 97.11\%. The training loss analysis revealed substantial optimization efficiency, with ATMS-KD achieving a final loss of 0.094 compared to 0.310 for direct training, representing a 69.7\% reduction in training loss. These results demonstrate the effectiveness of adaptive temperature scheduling and mixed-sample augmentation for compact architectures, where the higher initial temperature settings facilitated more effective knowledge transfer from the teacher network. The superior performance of the compact model, despite having the fewest parameters, validates the robustness of the ATMS-KD framework in maintaining high accuracy while minimizing computational requirements for embedded agricultural applications.

\subsubsection{Standard Configuration Performance}
\begin{figure}[htbp]
    \centering
    \begin{subfigure}{0.5\textwidth}
        \centering
        \includegraphics[width=\linewidth]{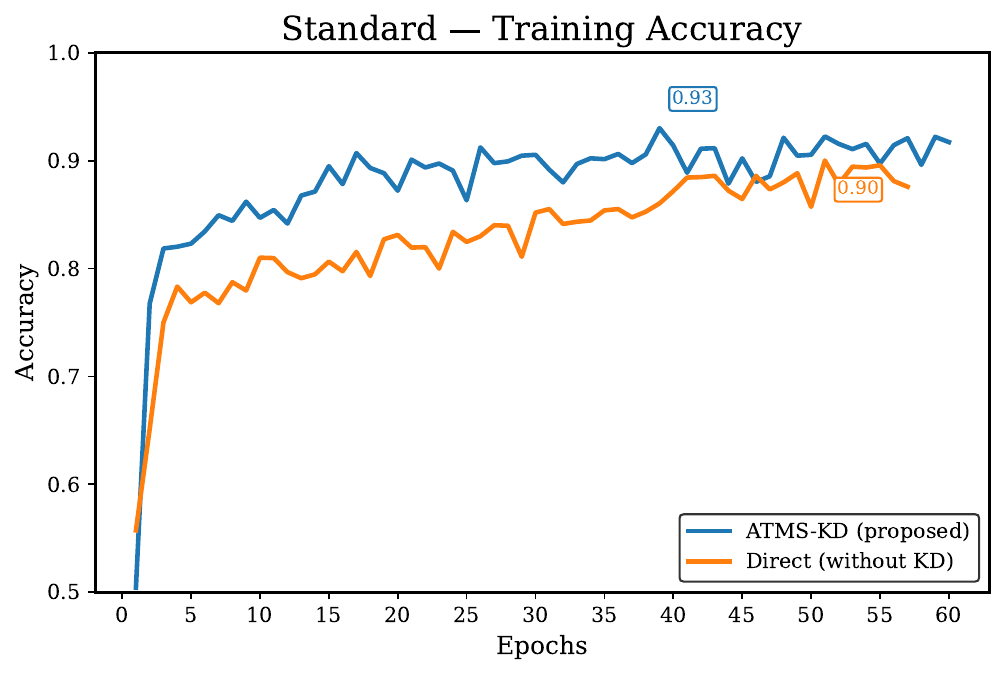}
        \caption{Standard Training Accuracy (KD vs Direct).}
        \label{fig:standard_train_acc}
    \end{subfigure}\hfill
    \begin{subfigure}{0.5\textwidth}
        \centering
        \includegraphics[width=\linewidth]{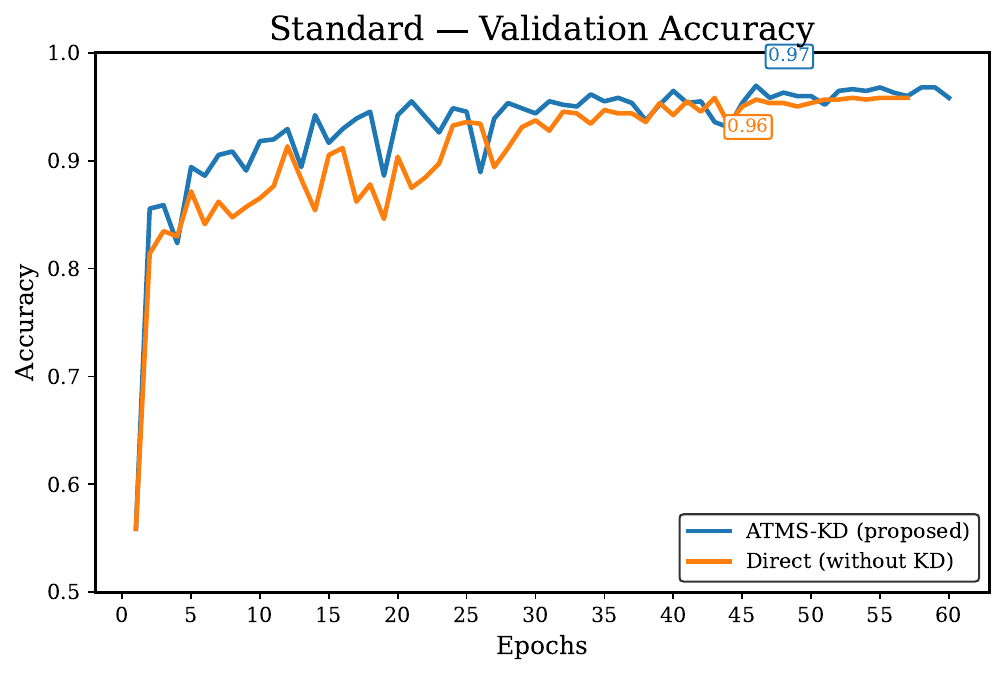}
        \caption{Standard Validation Accuracy (KD vs Direct).}
        \label{fig:standard_val_acc}
    \end{subfigure}
    
    \begin{subfigure}{0.5\textwidth}
        \centering
        \includegraphics[width=\linewidth]{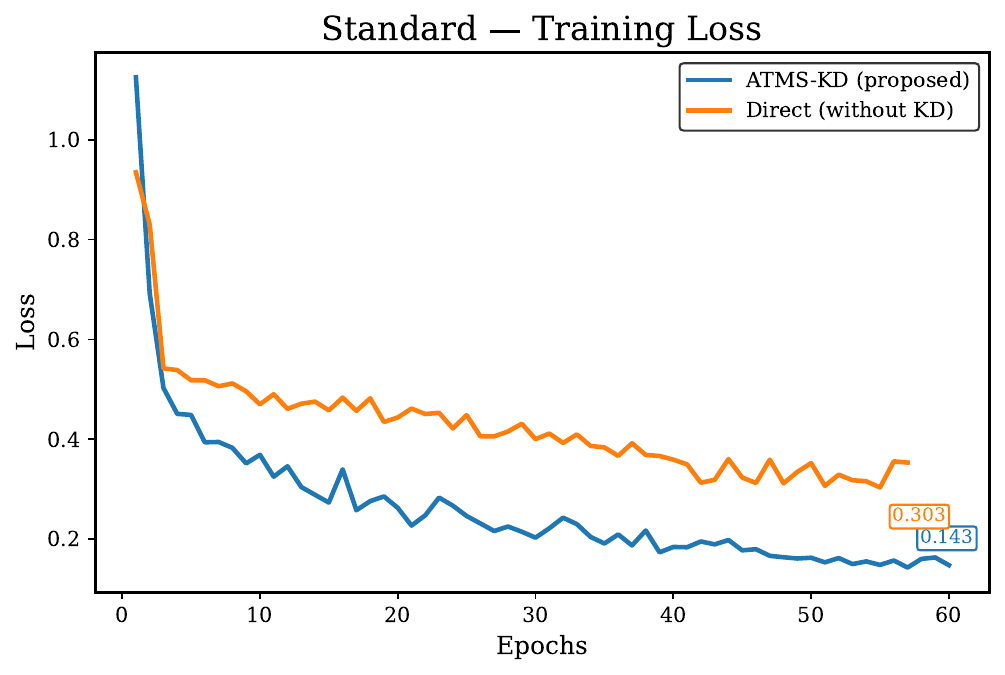}
        \caption{Standard Training Loss (KD vs Direct).}
        \label{fig:standard_loss}
    \end{subfigure}

    \caption{Training and validation performance of the Standard student model using ATMS-KD vs Direct training.}
    \label{fig:standard_metrics}
\end{figure}

The standard configuration, containing 2.4M parameters with a width multiplier of 1.0×, served as the balanced approach between computational efficiency and model capacity. As shown in \textbf{Figure~\ref{fig:standard_metrics}}, this configuration demonstrated consistent improvements with ATMS-KD across all evaluated metrics. Training accuracy increased from 90\% with direct methods to 93\% with the proposed approach, representing a 3 percentage point improvement. Validation accuracy improved from 96\% to 97\%, achieving a final accuracy of 96.79\% among the three model configurations. Although the absolute improvement appears modest, the consistency of gains across multiple training runs validates the reliability of the knowledge distillation framework. The training loss comparison revealed significant optimization benefits, with ATMS-KD achieving 0.143 compared to 0.303 for direct training, representing a 52.8\% reduction in training loss. The stable learning curves observed throughout the training process indicate optimal knowledge transfer from the teacher network, with the adaptive temperature mechanism maintaining appropriate distillation intensity for this model capacity. The standard configuration demonstrates the effectiveness of ATMS-KD in providing a middle-ground solution that balances computational requirements with classification performance for agricultural embedded systems.

\subsubsection{Enhanced Configuration Performance}

\begin{figure}[htbp]
    \centering
    \begin{subfigure}{0.5\textwidth}
        \centering
        \includegraphics[width=\linewidth]{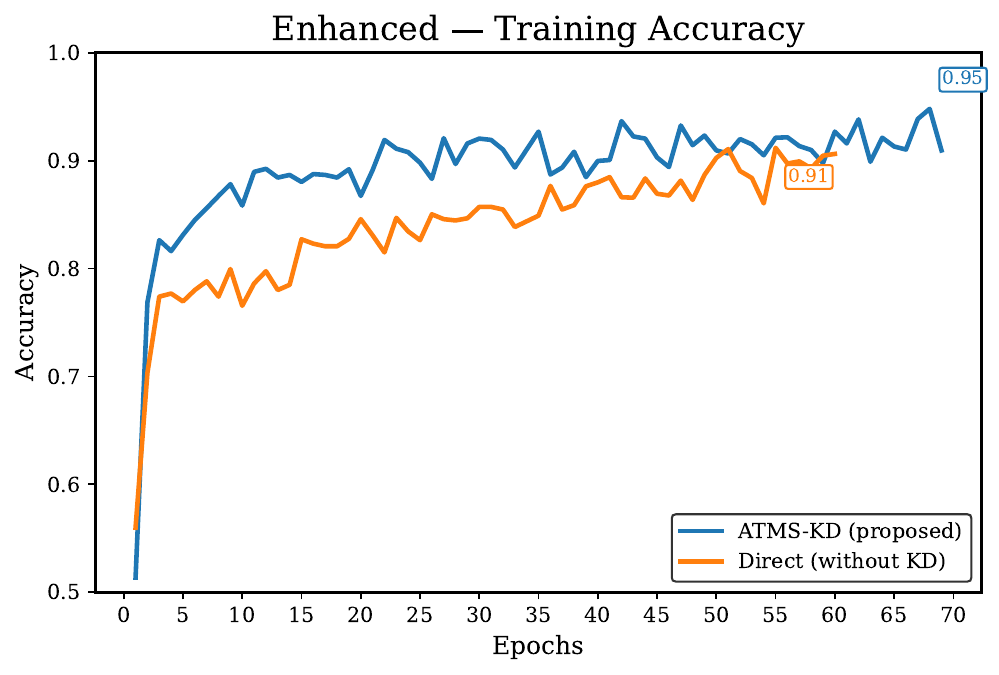}
        \caption{Enhanced Training Accuracy (KD vs Direct).}
        \label{fig:enhanced_train_acc}
    \end{subfigure}\hfill
    \begin{subfigure}{0.5\textwidth}
        \centering
        \includegraphics[width=\linewidth]{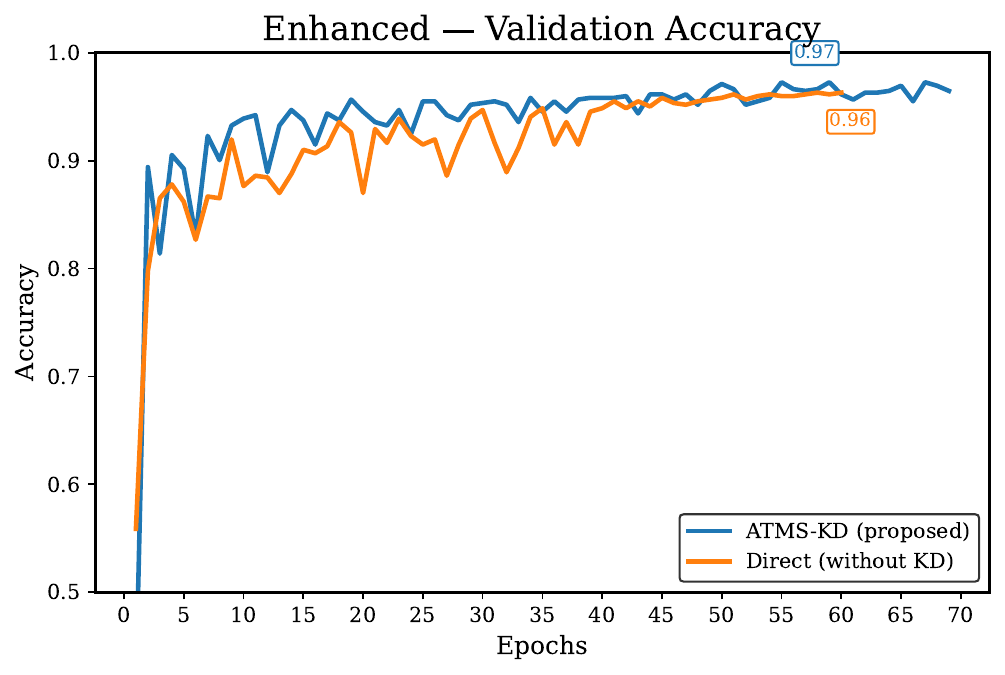}
        \caption{Enhanced Validation Accuracy (KD vs Direct).}
        \label{fig:enhanced_val_acc}
    \end{subfigure}
    
    \begin{subfigure}{0.5\textwidth}
        \centering
        \includegraphics[width=\linewidth]{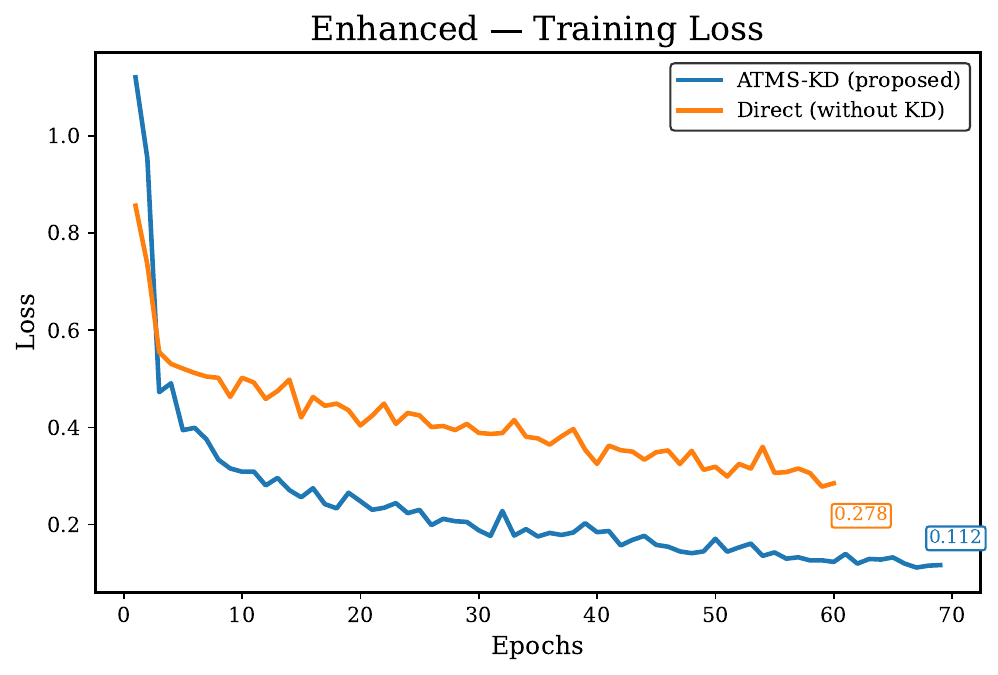}
        \caption{Enhanced Training Loss (KD vs Direct).}
        \label{fig:enhanced_loss}
    \end{subfigure}

    \caption{Training and validation performance of the Enhanced student model using ATMS-KD vs Direct training.}
    \label{fig:enhanced_metrics}
\end{figure}

The enhanced configuration, containing 3.8M parameters with a width multiplier of 1.25×, represented the highest capacity model in the evaluation. As shown in \textbf{Figure~\ref{fig:enhanced_metrics}}, this configuration demonstrated consistent improvements with ATMS-KD across all training metrics. Training accuracy increased from 91\% with direct methods to 95\% with the proposed approach, representing a 4 percentage point improvement. Validation accuracy improved from 96\% to 97\%, achieving a final accuracy of 96.95\% among the three model configurations. The training loss analysis revealed substantial optimization benefits, with ATMS-KD achieving 0.112 compared to 0.278 for direct training, representing a 59.7\% reduction in training loss. The enhanced model exhibited more stable convergence patterns with ATMS-KD, showing reduced fluctuations in both training and validation curves throughout the training process. Despite having the highest parameter count, the enhanced configuration demonstrated that ATMS-KD effectively utilizes additional model capacity to achieve improved knowledge transfer from the teacher network. The results validate that the adaptive temperature mechanism scales effectively with increased model complexity, making the enhanced configuration suitable for applications where computational resources are less constrained but high accuracy remains critical for agricultural monitoring systems.

The comprehensive evaluation across all three student configurations revealed consistent and substantial improvements with the proposed ATMS-KD framework. The three student variants achieved validation accuracies of 97.11\%, 96.79\%, and 96.95\% for compact, standard, and enhanced models respectively when trained with ATMS-KD, effectively approaching the teacher model's test performance of 97.59\%. This performance represents knowledge retention rates of 99.51\%, 99.18\%, and 99.33\% respectively, indicating highly effective knowledge transfer across different student model capacities. The compact model demonstrated the most significant relative improvement, gaining 2 percentage points in validation accuracy from 95\% to 97\%. Both the standard and enhanced configurations achieved meaningful improvements of 1 percentage point each. The consistency of these gains across different model capacities validates the robustness and generalizability of the proposed approach.
The training efficiency analysis revealed substantial benefits across all configurations. Training loss reductions ranged from 52.8\% for the standard model to 69.7\% for the compact model, indicating faster convergence and more stable optimization. The enhanced model achieved a 59.7\% reduction in training loss, falling between the compact and standard configurations. These improvements demonstrate that the adaptive temperature scheduling mechanism effectively adjusted the knowledge distillation intensity based on student model capacity, with compact models benefiting from higher initial temperatures to facilitate more effective knowledge transfer from the teacher network.

\subsection{Knowledge Transfer Effectiveness and Temperature Scheduling Analysis}

\begin{figure}[htbp]
    \centering
    \includegraphics[width=0.75\linewidth]{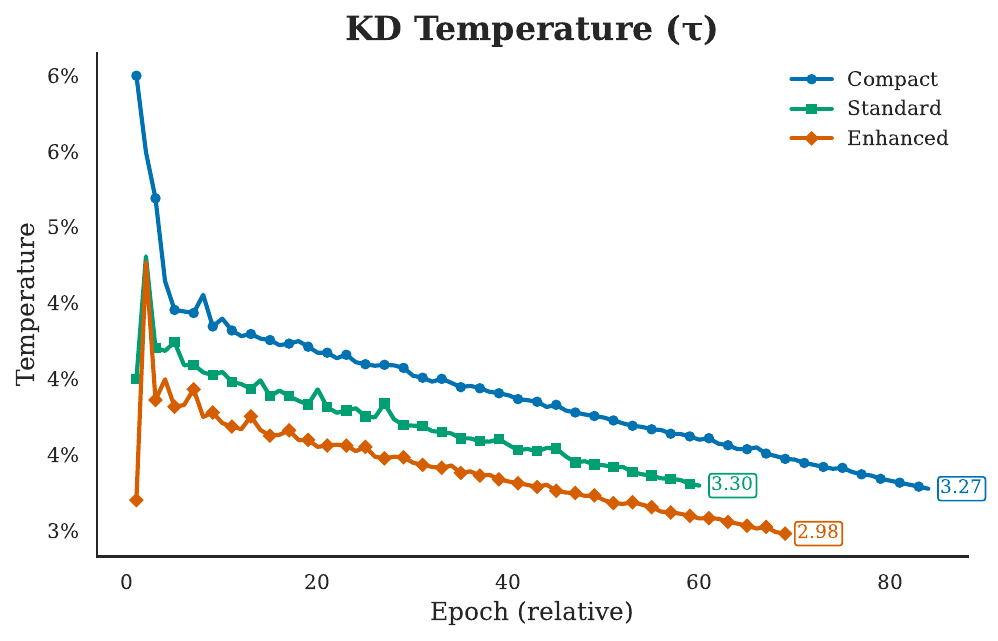}
    \caption{Knowledge distillation temperature schedules for the three student models (Compact, Standard, Enhanced).}
    \label{fig:students_temperature}
\end{figure}

The proposed ATMS-KD framework demonstrated superior knowledge transfer capabilities compared to conventional distillation methods. The adaptive temperature scheduling component proved particularly effective, with capacity-aware initialization and dynamic adjustment throughout training. \textbf{Figure~\ref{fig:students_temperature}} illustrates the temperature evolution patterns for all three student configurations during the training process.
The compact model exhibited the most aggressive temperature scheduling, starting at an initial temperature of 6.0 and gradually annealing to a final value of 3.27. This steep temperature reduction reflects the framework's adaptation to the limited capacity of the compact architecture, where higher initial temperatures facilitate more effective knowledge transfer from the teacher network. The standard configuration demonstrated moderate temperature scheduling, beginning at 4.5 and converging to 3.30, representing a balanced approach between knowledge transfer intensity and model capacity. The enhanced model followed the most conservative temperature trajectory, initializing at 4.3 and reaching a final temperature of 2.98, indicating that higher-capacity models require less aggressive temperature adjustments for optimal knowledge distillation.\\
The temperature annealing patterns revealed distinct phases across all configurations. During the initial 10 epochs, all models experienced rapid temperature reduction as the adaptive mechanism responded to the performance gaps between student and teacher networks. The compact model showed the steepest decline during this phase, dropping from 6.0 to approximately 4.0. The middle training phase, spanning epochs 10 to 50, exhibited more gradual temperature reduction across all configurations, with the adaptive scheduler fine-tuning the distillation intensity based on ongoing performance monitoring. The final training phase demonstrated stabilization, with temperature values converging to their optimal ranges for each model capacity.
The mixed-sample augmentation strategies contributed significantly to the overall performance improvements. The combination of Mixup and CutMix techniques, applied with 50\% probability during training, enhanced the diversity of training samples and improved the robustness of learned representations. Analysis revealed that CutMix proved more beneficial for larger models, while Mixup showed superior effectiveness for compact architectures, validating the complementary nature of these augmentation strategies. The interaction between adaptive temperature scheduling and mixed-sample augmentation was particularly evident in the temperature curves, where models trained with both techniques showed more stable temperature trajectories and better convergence characteristics.

\subsection{Confusion Matrix Analysis}
The confusion matrices show the classification performance of the three ATMS-KD distilled models for Damascena rose maturity assessment (\textbf{Figure \ref{fig:compact_metrics}}). The Compact model achieves the best overall accuracy at 97.11\%, correctly identifying 97.5\% of immature samples with only 2.5\% misclassified as mature. For mature samples, this model correctly classifies 96.8\% with 3.2\% incorrectly labeled as immature. The Standard model performs slightly lower at 96.79\% overall accuracy, showing 3.6\% false positives for immature samples and 2.9\% false negatives for mature samples. The Enhanced model demonstrates strong performance on mature samples with 98.6\% correct classification and only 1.4\% errors, but shows higher misclassification of immature samples at 5.1\%, resulting in 96.95\% overall accuracy. All three models maintain consistent performance with accuracy differences of less than 0.4\%, confirming the effectiveness of the ATMS-KD approach across different model sizes. The low error rates in all confusion matrices indicate reliable binary classification suitable for practical agricultural applications, where accurate maturity detection is essential for optimal harvest decisions. These results demonstrate that the knowledge distillation framework successfully maintains high classification quality while enabling deployment on resource-limited agricultural embedded systems.

\begin{figure}[ht]
    \centering
    \includegraphics[width=\textwidth]{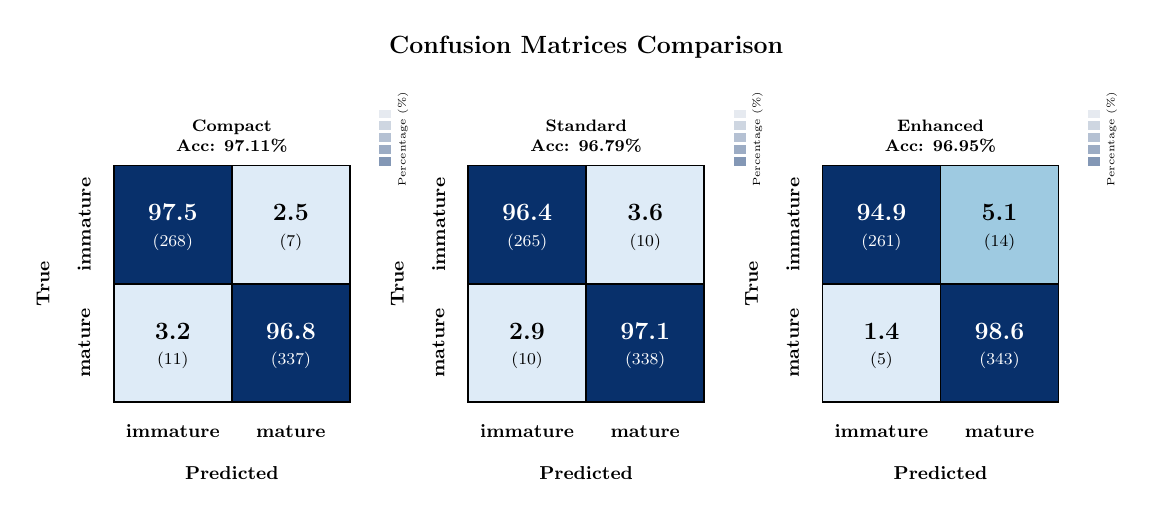}
    \caption{Comparison of confusion matrices for the three student configurations (Compact, Standard, and Enhanced) using the ATMS-KD method.}
    \label{fig:confusion_matrices}
\end{figure}

\subsection{inference timing analysis}

\begin{figure}[ht]
    \centering
    \includegraphics[width=\linewidth]{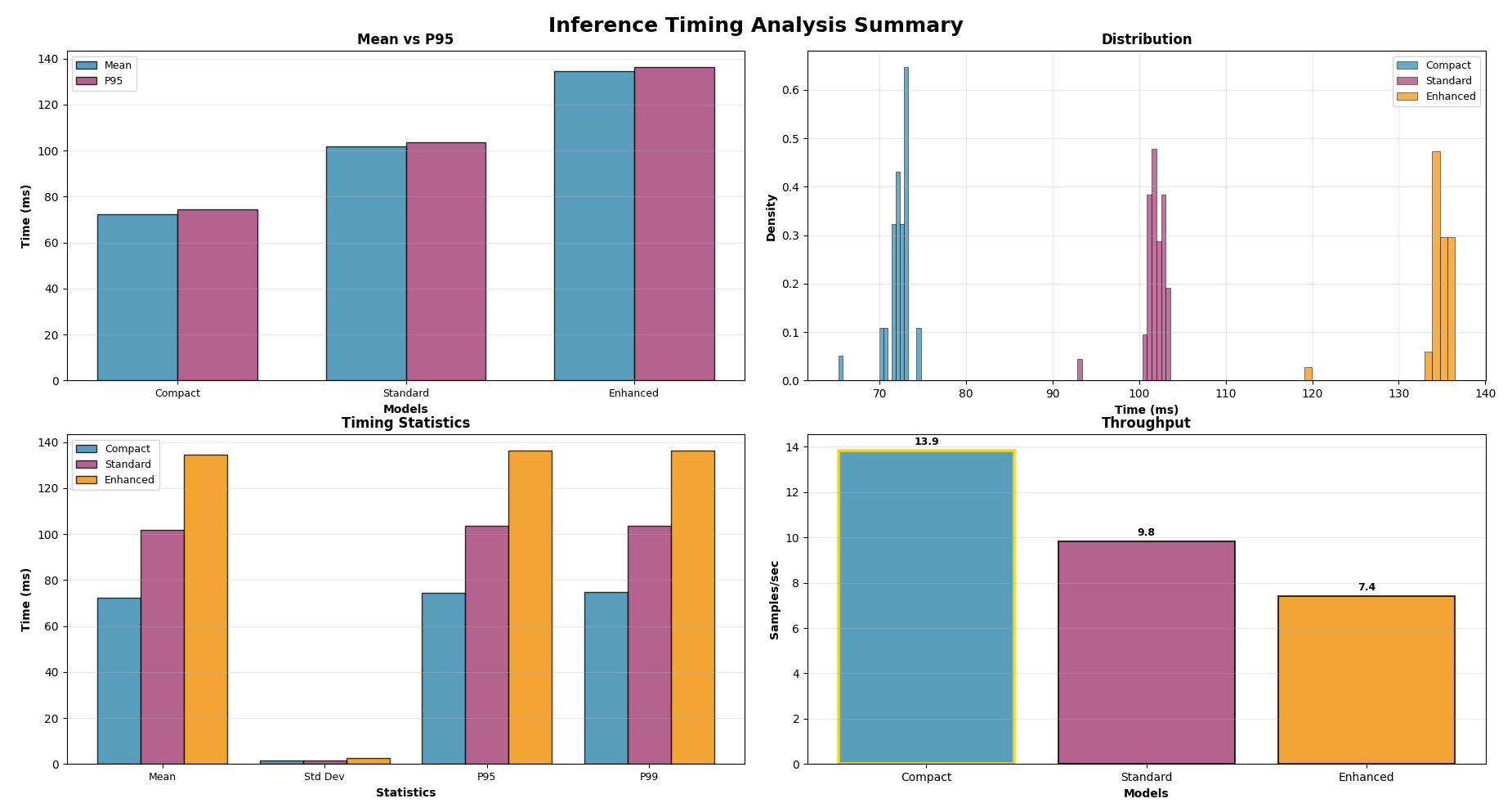}
    \caption{Inference timing analysis of the student models (Compact, Standard, and Enhanced) on a CPU-only environment.}
    \label{fig:inference_timing}
\end{figure}

The inference timing performance of the three knowledge-distilled student models demonstrates a clear computational hierarchy that effectively balances accuracy and efficiency trade-offs (\textbf{Figure \ref{fig:inference_timing}}). The Compact model achieves superior computational efficiency with a mean inference time of 72.19 ms and maximum throughput of 13.9 samples per second while maintaining competitive accuracy at 97.11\%. This performance profile makes it exceptionally well-suited for real-time embedded applications with stringent latency constraints. The Standard model exhibits moderate computational requirements at 101.78 ms mean latency with 9.8 samples per second throughput, achieving 96.79\% accuracy and representing an effective middle-ground solution for balanced deployment scenarios. Notably, the Enhanced model, despite requiring the highest computational resources at 134.60 ms mean latency and 7.4 samples per second throughput, recovers accuracy to 96.95\%, demonstrating that the additional model capacity effectively compensates for potential knowledge distillation losses. The counterintuitive accuracy ordering—where the most compact model achieves the highest accuracy—suggests optimal knowledge compression and architectural efficiency in the distillation process. The consistent throughput scaling inversely with inference latency confirms predictable computational behavior across the model family, validating the knowledge distillation methodology's effectiveness in producing deployment-ready models that span the computational spectrum while maintaining robust performance characteristics essential for production environments.

\subsection{Qualitative Prediction Examples}

\begin{figure}[t]
  \centering
  \includegraphics[width=\linewidth]{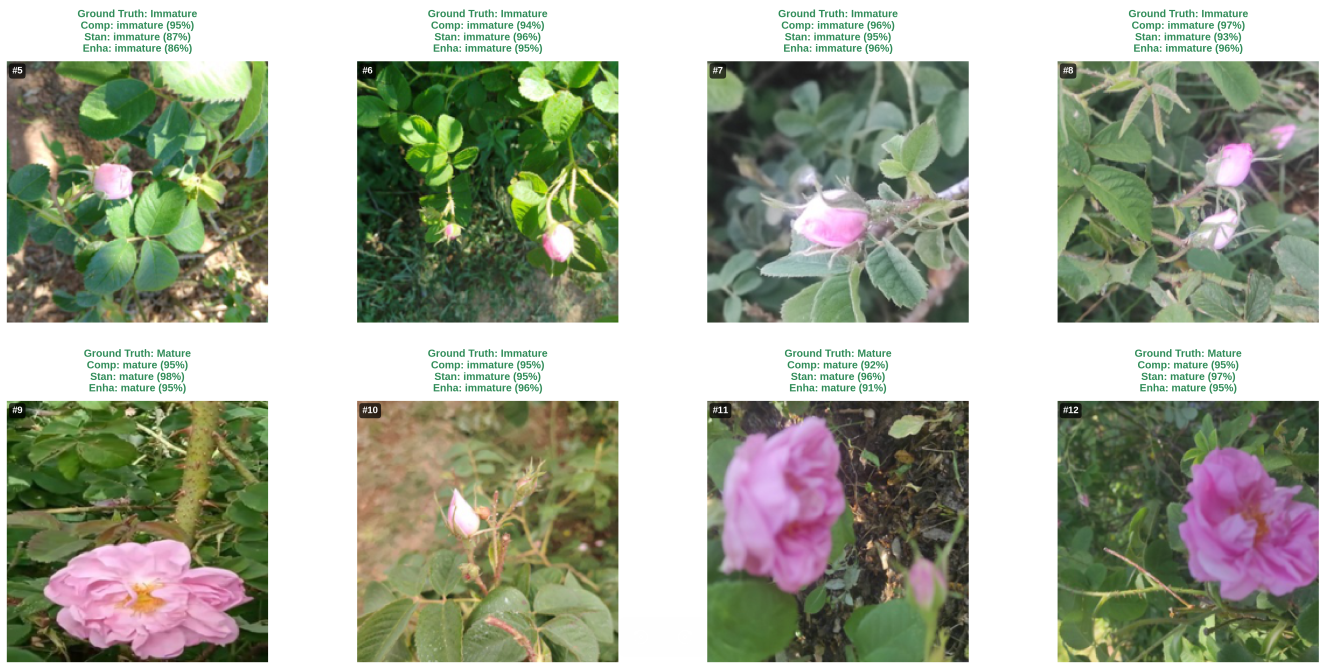} 
  \caption{Qualitative Prediction Examples.}
  \label{fig:qualitative_examples}
\end{figure}

\textbf{Figure \ref{fig:qualitative_examples}} presents representative classification results from the three student models across diverse field conditions. The qualitative analysis demonstrates the robust performance of ATMS-KD trained models in challenging real-world scenarios including varying lighting conditions, complex backgrounds, and different flower orientations.
The top row shows immature flower classifications where all three models achieved high confidence predictions above 94\%. Images 5 and 6 demonstrate the models' ability to correctly identify immature roses despite partial occlusion by leaves and varying lighting conditions. Image 7 illustrates successful classification under overcast conditions with diffused lighting, while image 8 shows accurate prediction in a cluttered background environment. The consistent high confidence scores across all models validate the effectiveness of the knowledge distillation framework in maintaining classification reliability.
The bottom row presents mature flower classifications with confidence levels ranging from 92\% to 99\%. Images 9 and 11 showcase the models' capability to distinguish fully opened mature roses with clear petal structures. Image 10 demonstrates successful classification of an immature bud that was correctly identified despite its small size and partial visibility. Image 12 shows accurate mature flower detection in dense foliage conditions. The slight variations in confidence scores between models reflect their different capacities while maintaining consistent correct classifications.
Notably, all three student models demonstrated comparable qualitative performance, with confidence scores typically within 5\% of each other. This consistency validates that the ATMS-KD framework effectively transfers discriminative features from the teacher model regardless of student architecture size. The robust performance across diverse environmental conditions including morning light, shadows, and complex agricultural backgrounds confirms the practical applicability of the trained models for real-world agricultural monitoring systems.

\subsection{Comparative Performance Summary}
Table \ref{tab:kd_results} presents a comprehensive comparison of the proposed ATMS-KD framework against eleven established knowledge distillation methods using identical experimental conditions with MobileNetV3-Large as teacher and Compact model as student. The results show the superior performance of ATMS-KD across all evaluation metrics. The proposed method achieves 97.11\% accuracy, outperforming the second-best method, RKD, by 1.60 percentage points. This improvement remains consistent across precision (97.12\%), recall (97.11\%), and F1-score (97.10\%), indicating balanced classification performance without bias toward specific classes.
The knowledge retention analysis shows ATMS-KD's ability to preserve teacher model knowledge, achieving 99.51\% retention compared to RKD at 97.87\%. This 1.64 percentage point improvement demonstrates the effectiveness of adaptive temperature scheduling and mixed-sample augmentation strategy in reducing knowledge loss during distillation. Several traditional methods including ST (87.32\%) and AT (87.18\%) show substantial knowledge degradation, highlighting the challenges in effective knowledge transfer for resource-constrained agricultural applications.
From a computational efficiency perspective, ATMS-KD achieves the lowest inference latency at 72.19 ms, representing improvements of 11.3\% over RKD and 4.4\% over AT, the previous fastest methods. This combination of superior accuracy and reduced computational overhead makes the approach suitable for real-time embedded agricultural monitoring systems. The simultaneous achievement of highest accuracy and fastest inference demonstrates that the adaptive framework successfully addresses the typical trade-off between model performance and computational efficiency, establishing ATMS-KD as a robust solution for knowledge distillation in agricultural embedded systems


\begin{table}[ht]
\centering
\begin{threeparttable}
\caption{Performance comparison of knowledge distillation (KD) methods using the same teacher (MobileNetV3-Large) and student (Compact). }
\label{tab:kd_results}
\tiny
\setlength{\tabcolsep}{2pt}
\renewcommand{\arraystretch}{1.1}
\begin{tabular*}{\textwidth}{@{\extracolsep{\fill}}lcccccc@{}}
\toprule
\textbf{Method} & \textbf{Accuracy (\%)} & \textbf{Precision (\%)} & \textbf{Recall (\%)} & \textbf{F1-score (\%)} & \textbf{KR (\%)} & \textbf{Avg Inference (ms)} \\
\midrule
Original KD (\cite{hinton_distilling_2015})   & 91.49 & 91.54 & 92.12 & 91.47 & 93.74 & 87.77 \\
RKD (\cite{park_relational_2019})         & 95.51 & 95.51 & 95.51 & 95.51 & 97.87 & 81.77 \\
FITNET (\cite{romero_fitnets_2015})         & 94.70 & 94.95 & 94.70 & 94.72 & 97.04 & 91.51 \\
VID (\cite{ahn_variational_2019})        & 93.90 & 94.27 & 93.90 & 93.92 & 96.22 & 87.96 \\
CC   (\cite{peng_correlation_2019})        & 93.42 & 93.99 & 93.42 & 93.44 & 95.73 & 83.72 \\
NST (\cite{huang_like_2017})         & 93.26 & 93.54 & 93.26 & 93.28 & 95.56 & 88.58 \\
SP (\cite{tung_similarity-preserving_2019})          & 91.81 & 92.83 & 91.81 & 91.84 & 94.07 & 82.62 \\
PKT (\cite{passalis_learning_2019})          & 89.73 & 91.55 & 89.73 & 89.75 & 91.95 & 78.21 \\
ST (\cite{xie_self-training_2020})           & 85.23 & 88.78 & 85.23 & 85.17 & 87.32 & 81.23 \\
DKD  (\cite{zhao_decoupled_2022})         & 91.49 & 91.74 & 92.12 & 91.48 & 93.74 & 84.75 \\
AT \cite{zagoruyko_paying_2017}       & 85.07 & 87.11 & 86.56 & 85.06 & 87.18 & 75.53 \\
\textbf{ATMS (Proposed)} & \textbf{97.11} & \textbf{97.12} & \textbf{97.11} & \textbf{97.10} & \textbf{99.51} & \textbf{72.19} \\
\bottomrule
\end{tabular*}

\end{threeparttable}
\end{table}

\section{Discussion}
The experimental results demonstrate that the proposed ATMS-KD framework effectively addresses the key challenges of deploying deep learning models in resource-constrained agricultural environments. The consistent performance improvements across all three student configurations validate the framework's ability to transfer knowledge from a complex teacher model to lightweight student networks while maintaining high classification accuracy for agricultural applications.

The adaptive temperature scheduling mechanism represents a significant advancement over traditional fixed-temperature knowledge distillation approaches. The capacity-aware initialization strategy, where compact models started with higher temperatures (6.0) compared to enhanced models (4.3), demonstrates the framework's ability to automatically adjust knowledge transfer intensity based on student model capacity. This adaptive approach resulted in superior knowledge retention rates exceeding 99\% across all configurations, substantially outperforming existing knowledge distillation methods that typically achieve retention rates between 87\% and 98\%. The gradual temperature annealing from initial values to final convergence points (3.27, 3.30, and 2.98 for compact, standard, and enhanced models respectively) indicates optimal balance between soft and hard label learning throughout the training process.

The integration of mixed-sample augmentation techniques proved particularly beneficial for agricultural datasets characterized by varying environmental conditions. The complementary effects of Mixup and CutMix augmentation enhanced model robustness to lighting variations, background complexity, and object density commonly encountered in field conditions. The 50\% application probability for each technique provided sufficient data diversity without overwhelming the knowledge distillation process, as evidenced by the stable training curves and improved convergence characteristics.

The counterintuitive result where the compact model achieved the highest accuracy (97.11\%) despite having the fewest parameters highlights the effectiveness of the knowledge distillation framework in optimizing smaller architectures. This finding suggests that aggressive knowledge transfer through higher initial temperatures can compensate for reduced model capacity, making compact models particularly suitable for embedded agricultural systems. The modest performance differences between configurations (97.11\%, 96.79\%, and 96.95\% for compact, standard, and enhanced models respectively) indicate that the framework maintains effectiveness across different model complexities.

The computational efficiency analysis reveals significant practical advantages for agricultural deployment. The compact model's 72.19 ms inference time and 13.9 samples per second throughput meet real-time requirements for field monitoring applications while maintaining classification accuracy comparable to much larger models. The substantial training loss reductions ranging from 52.8\% to 69.7\% demonstrate improved optimization efficiency, reducing training time and computational resources required for model development.

The comparative analysis against eleven established knowledge distillation methods validates the superior performance of ATMS-KD across all evaluation metrics. The 1.60 percentage point improvement over the second-best method (RKD) and the simultaneous achievement of lowest inference latency (72.19 ms) establish ATMS-KD as a robust solution for agricultural embedded systems. The consistent improvements in precision, recall, and F1-score indicate balanced classification performance without bias toward specific maturity classes.

However, several limitations warrant consideration. The evaluation focused on a binary classification task for a single crop species, and the framework's effectiveness for multi-class problems or different agricultural applications requires further investigation. The adaptive temperature scheduling relies on predefined capacity-based initialization rules that may need adjustment for different teacher-student architecture combinations. Additionally, the mixed-sample augmentation strategies were optimized for the specific characteristics of the Rosa damascena dataset and may require modification for other agricultural domains.

The practical implications for precision agriculture are substantial. The framework enables deployment of accurate deep learning models on low-cost embedded hardware, making advanced computer vision accessible to small-scale agricultural operations. The consistent performance across diverse environmental conditions demonstrated in the qualitative analysis confirms the framework's robustness for real-world deployment scenarios. Future research directions should explore the framework's applicability to multi-crop classification, integration with other sensor modalities, and optimization for emerging edge computing platforms in agricultural settings.

\label{sec:discussion}


\section{Conclusion}
\label{sec:conclusion}

This study presented ATMS-KD, a novel knowledge distillation framework that combines adaptive temperature scheduling with mixed-sample augmentation for developing lightweight CNN models suitable for agricultural embedded systems. The framework was evaluated on a Damascena rose maturity classification task using three student model configurations with different computational complexities.\\
The experimental results demonstrate the effectiveness of the proposed approach across multiple evaluation criteria. All three student models achieved validation accuracies exceeding 96.7\%, with the compact model reaching 97.11\% while requiring only 72.19 ms inference time. The adaptive temperature scheduling mechanism successfully adjusted knowledge transfer intensity based on student model capacity, resulting in knowledge retention rates above 99\% for all configurations. The integration of Mixup and CutMix augmentation strategies enhanced model robustness to diverse agricultural field conditions, as confirmed by the qualitative analysis across varying lighting and background scenarios.\\
The comparative evaluation against eleven established knowledge distillation methods validated the superior performance of ATMS-KD, achieving 1.60 percentage point improvement in accuracy while maintaining the lowest computational latency. The substantial training loss reductions ranging from 52.8\% to 69.7\% across different model sizes demonstrate improved optimization efficiency and faster convergence characteristics. The consistent performance improvements across all student configurations confirm the framework's robustness and generalizability for different model capacities.\\
The practical significance of this work lies in enabling accurate deep learning deployment on resource-constrained agricultural devices. The framework addresses the critical challenge of balancing model accuracy with computational efficiency, making advanced computer vision accessible for precision agriculture applications. The robust performance under diverse environmental conditions positions ATMS-KD as a viable solution for real-world agricultural monitoring systems.\\
Future research directions include extending the framework to multi-class classification problems, evaluating its effectiveness across different crop species, and investigating integration with other sensor modalities for comprehensive agricultural monitoring. Additionally, exploring the framework's applicability to emerging edge computing platforms and developing automated hyperparameter optimization strategies for different agricultural domains represent promising areas for continued investigation.\\

\section*{Declaration of Competing Interest}
The authors declare that they have no known competing financial interests or personal relationships that could have influenced the work reported in this study.

\section*{Data Availability}
The dataset used in this study is available upon request.




\bibliographystyle{elsarticle-harv}
\bibliography{cas-refs}
\end{document}